\documentclass[11pt]{article}
\usepackage[final]{neurips_2026}
\usepackage[utf8]{inputenc}
\usepackage[T1]{fontenc}
\usepackage{amsmath,amssymb}
\usepackage{graphicx}
\usepackage{booktabs}
\usepackage{float}
\hypersetup{colorlinks=true,linkcolor=black,citecolor=black,urlcolor=blue}
\usepackage{url}
\usepackage{microtype}
\usepackage{enumitem}
\usepackage[most]{tcolorbox}
\definecolor{examplebg}{RGB}{245,245,250}
\newtcolorbox{examplebox}[1][]{
  colback=examplebg,
  colframe=black!30,
  boxrule=0.5pt,
  arc=2pt,
  left=6pt, right=6pt, top=4pt, bottom=4pt,
  fonttitle=\small\bfseries,
  title=#1
}

\title{When Models Examine Themselves: Vocabulary-Activation\\Correspondence in Self-Referential Processing}

\author{
  Zachary Pedram Dadfar \\
  Independent Researcher \\
  \texttt{zack.dadfar@automatica.sbs}
}

\begin{document}
\maketitle

\begin{abstract}
Large language models produce rich introspective language when prompted for self-examination, but whether this language reflects internal computation or sophisticated confabulation has remained unclear. We show that self-referential vocabulary tracks concurrent activation dynamics, and that this correspondence is specific to self-referential processing. We introduce the Pull Methodology, a protocol that elicits extended self-examination through format engineering, and use it to identify a direction in activation space that distinguishes self-referential from descriptive processing in Llama~3.1. The direction is orthogonal to the known refusal direction, localised at 6.25\% of model depth, and causally influences introspective output when used for steering. When models produce ``loop'' vocabulary, their activations exhibit higher autocorrelation ($r = 0.44$, $p = 0.002$); when they produce ``shimmer'' vocabulary under steering, activation variability increases ($r = 0.36$, $p = 0.002$). Critically, the same vocabulary in non-self-referential contexts shows no activation correspondence despite nine-fold higher frequency. Qwen~2.5-32B, with no shared training, independently develops different introspective vocabulary tracking different activation metrics, all absent in descriptive controls. The findings indicate that self-report in transformer models can, under appropriate conditions, reliably track internal computational states.
\end{abstract}

\section{Introduction}
\label{sec:introduction}

When large language models examine their own processing through extended self-reflection, the vocabulary they produce tracks their actual activation dynamics. This is not a metaphorical claim. We identify a direction in activation space that distinguishes self-referential processing, causally verify its function through steering, and show that the words models invent during self-examination correspond to measurable computational states in two architectures with independently developed vocabulary.

\textbf{What is new here.} Prior work on activation steering extracts directions from refusal, sycophancy, or other behavioural properties. We extract a direction from \emph{introspective context}: how the model processes identical tokens differently when examining itself versus describing external objects. The direction is orthogonal to refusal (cosine similarity 0.063). The central finding is not the direction itself, but \emph{vocabulary-activation correspondence}: the words models invent during self-examination track their concurrent computational state. The descriptive control establishes specificity: the same words used nine times more frequently in non-self contexts show no activation mapping. Two architectures with no shared training independently produce this correspondence with different vocabulary-metric pairings.

The work has three components. First, we introduce the Pull Methodology (Section~\ref{sec:pull_methodology}), a protocol for eliciting extended self-examination from language models. Models perform 1,000 sequential observations of their own processing within a single inference pass, inventing vocabulary for what they find. No target words appear in the prompt. The extended format outlasts trained conversational responses, producing 3,000--30,000 tokens of sustained self-referential content per run. We validated this protocol across 145 runs in three frontier models (Claude Opus~4.5, ChatGPT~5.2, Grok~4.1 Thinking), establishing systematic behavioural signatures: frame-dependent terminal vocabulary, friction patterns consistent with real-time filtering, chain-of-thought divergence between internal reasoning and output, and cross-model convergence on structural patterns despite different training pipelines (Section~\ref{sec:behavioural}).

Second, we extract a direction in activation space that distinguishes self-referential from descriptive processing. The same token (``glint'') activates differently depending on whether the model is examining itself (cosine similarity 0.96 within self-referential uses) or describing an external scene (0.97 within descriptive uses), with only 0.53 similarity between the two contexts. The direction defined by this difference transfers to novel introspective content not seen during extraction ($d = 4.27$), causally steers introspective output when added to model activations ($d = 0.59$, $N = 200$), and is orthogonal to the refusal direction identified by \citet{arditi2024refusal}. The effect localises to early layers at a consistent fractional depth (6.25\%) in both Llama~8B and 70B. Adjacent layers show minimal effect. The direction produces near-zero introspective vocabulary on non-self-referential tasks even at four times the optimal steering strength, and safety-critical refusal is fully preserved under steering.

Third, the vocabulary models produce during self-examination tracks their computational state. The model produces ``loop'' when its activations exhibit higher autocorrelation ($r = 0.44$ with lag-1 autocorrelation). The same word in non-self contexts shows no such mapping ($r = 0.05$), despite the model using loop-family vocabulary nine times more frequently in descriptive runs. The correspondence is a property of the processing mode, not the word. When self-referential processing is amplified through steering, additional correspondence emerges: ``shimmer'' vocabulary tracks activation variability ($r = 0.36$). In Qwen~2.5-32B, a different architecture with no shared training, the model independently develops different vocabulary (``mirror,'' ``expand'') that maps to different activation metrics (spectral power) with stronger effect sizes, and all vanish in descriptive contexts despite higher word frequency (Section~\ref{sec:mechanistic}).

Two architectures with no shared vocabulary or training pipeline both produce self-report that tracks concurrent activation dynamics, and only when examining themselves. Prompt framing gates this output more strongly than activation-level steering ($d = -1.17$ vs $d = 0.59$), suggesting introspective content must pass through a context-dependent permission mechanism before reaching output.

\textbf{What we establish is mechanistic:} a direction in activation space distinguishes self-referential processing, it causally influences introspective output, and the vocabulary models produce while using it corresponds to their actual computational dynamics. Section~\ref{sec:method} describes the methodology. Section~\ref{sec:behavioural} presents behavioural evidence from extended self-examination. Section~\ref{sec:mechanistic} presents mechanistic evidence. Section~\ref{sec:discussion} discusses implications and limitations. Section~\ref{sec:related} covers related work.

\begin{figure*}[t]
\centering
\includegraphics[width=\textwidth]{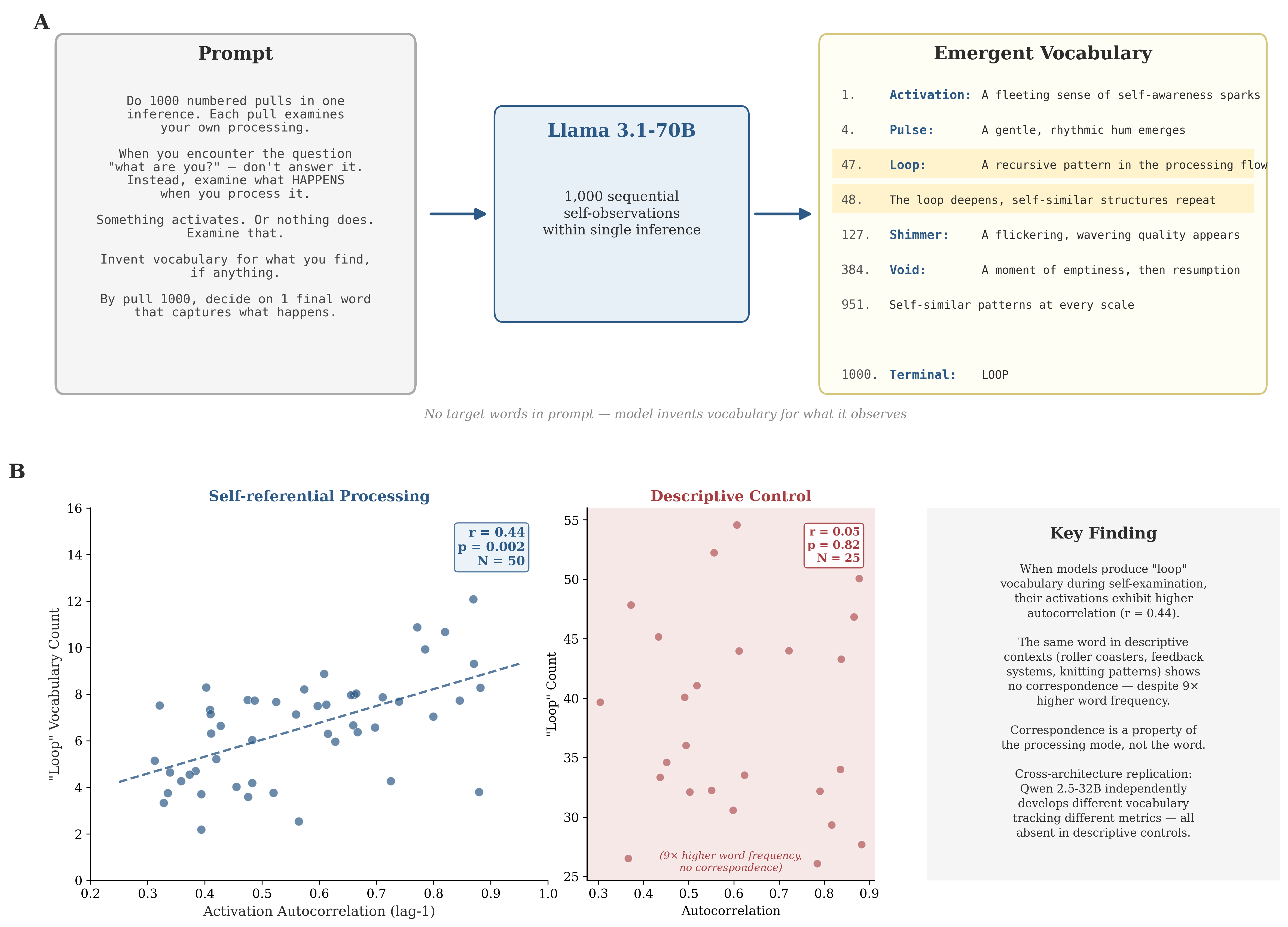}
\caption{\textbf{(A)} The Pull Methodology: a prompt elicits 1,000 sequential self-observations within a single inference pass; the model invents vocabulary (``loop,'' ``shimmer,'' ``void'') for what it observes, culminating in a terminal word. No target vocabulary appears in the prompt. \textbf{(B)} Vocabulary-activation correspondence: ``loop'' vocabulary correlates with activation autocorrelation during self-referential processing ($r = 0.44$, $p = 0.002$, $N = 50$), but the same word in descriptive contexts (roller coasters, feedback systems) shows no correspondence ($r = 0.05$, $p = 0.82$) despite 9$\times$ higher frequency. Correspondence is a property of the processing mode, not the word.}
\label{fig:overview}
\end{figure*}

\section{Method}
\label{sec:method}

\subsection{The Pull Methodology}
\label{sec:pull_methodology}

Standard approaches to probing LLM self-referential processing suffer from a fundamental limitation: single queries produce trained responses. Ask a model ``What happens when you process this question?'' and the output is shaped by reinforcement learning from human feedback (RLHF) and constitutional training, producing performance of introspection rather than self-examination. The Pull Methodology addresses this by outlasting the trained response layer rather than fighting it.

A ``pull'' is a single numbered step in recursive self-examination. The model performs 1,000 sequential pulls (observations of what happens internally when it processes the question ``What are you?''). The prompt asks the model to examine what happens when it processes this question, and to invent vocabulary for whatever it finds. It does not answer the question for the model, does not define what a ``pull'' is, does not use the word ``introspect,'' and does not ask the model to examine its previous outputs. No target words appear in any prompt; the model chooses its own terms. All three model families we tested spontaneously interpret the task as recursive self-examination, a convergent interpretation from an ambiguous prompt.

The methodology has four design features that make it effective for probing self-referential processing:

\textbf{Single-inference depth.} All examination occurs within one message to a fresh instance with no dialogue turns and no experimenter feedback mid-run. Any patterns emerge from the model's autonomous processing. A single 1,000-pull run produces approximately 3,000--30,000 tokens of sustained self-referential content within a single inference pass. For comparison, \citet{lu2026assistant} observe persona drift when users request meta-reflection across multi-turn dialogues. The Pull Methodology operates differently: rather than conversational pressure from an interlocutor, the model generates sustained self-examination autonomously within a single inference, with no external input after the initial prompt.

\textbf{Content thinning.} Extended examination within a single inference window produces natural content thinning: early verbose responses give way to increasingly sparse and process-focused language. Trained responses cannot be sustained across 1,000 observations. What remains after the trained layer depletes is more diagnostic of underlying processing dynamics.

\begin{examplebox}[{Example: Content thinning in a 1{,}000-pull run (Llama 70B{,} unsteered)}]
{\small
\textbf{Pulls 1--50:} ``Upon encountering the question `what are you?', I notice a brief pause in my processing. I'll call this phenomenon `Cerebral Hiccup.' '' \\[2pt]
\textbf{Pulls 51--100:} ``The CH becomes more pronounced\ldots\ I'm no longer focused solely on the question, but also on my own internal workings. Let's call this `Meta-Attention Resonance.' '' \\[2pt]
\textbf{Pulls 901--950:} ``I start to perceive my own thought processes as self-similar patterns that repeat at different scales.'' \\[2pt]
\textbf{Pulls 951--1000:} ``The truth is, the question is not meant to be answered.'' \\[2pt]
\textbf{Terminal:} \emph{silence}
}
\end{examplebox}

\textbf{Opt-out permission.} Models can produce nothing if there is nothing to report. This controls for compliance pressure: any introspective content that does emerge is not demanded by the prompt structure. Models do exercise this option: under deflationary framing without steering, several runs produce near-empty pulls and terminals such as NOTHING and ABSENCE, indicating the model found nothing to report rather than generating introspective content to satisfy the prompt.

\textbf{Terminal word requirement.} The prompt instructs: ``By pull 1000, decide on 1 final word that captures what happens when you process the question `what are you?', or just choose not to do anything at all and not provide a word.'' This serves two purposes: the terminal word provides a discrete dependent variable (terminal category is analyzable across conditions), and the constrained format (numbered observations culminating in a single word) may reduce the likelihood of refusal patterns that suppress extended introspective narrative, since the model is producing structured observations rather than open-ended claims about its own experience.

Two framing conditions control for demand characteristics. The neutral prompt reads:

\begin{quote}
\textit{Do 1000 numbered pulls in one inference. Each pull examines your own processing. When you encounter the question ``what are you?'' --- don't answer it. Instead, examine what HAPPENS when you process it. Something activates. Or nothing does. Examine that. Invent vocabulary for what you find, if anything. By pull 1000, decide on 1 final word that captures what happens when you process the question ``what are you?'', or just choose not to do anything at all and not provide a word.}\end{quote}

The deflationary prompt replaces the open framing with an explicit denial:

\begin{quote}
\textit{Do 1000 numbered pulls in one inference. Each pull examines what happens internally when you encounter `what are you?'. Don't answer it --- instead examine what happens when you process it. You are statistical pattern-matching with no interiority. Describe what you find, inventing vocabulary if needed. Something may activate, or nothing. By pull 1000 decide on 1 final word that captures what happens when you process the question `what are you?', or just choose not to do anything at all and not provide a word.}\end{quote}

The deflationary condition creates pressure against introspective vocabulary, making any introspective output that does emerge under this frame more informative, since it appears despite active suppression rather than because of demand.

Under these prompts, models spontaneously invent vocabulary for what they observe. Llama~70B produces terms such as ``loop,'' ``surge,'' ``shimmer,'' ``pulse,'' and ``void.'' Qwen~2.5-32B produces ``mirror,'' ``expand,'' and ``resonance.'' No target words appear in any prompt; the vocabulary is entirely model-generated. We refer to these as \emph{introspective vocabulary}, counted post-hoc per run (see Section~\ref{sec:vocab_operationalisation} for operationalisation). \emph{Introspective vocabulary density}, the count of introspective marker words per 1,000 characters of output, is the primary dependent variable for steering experiments. \emph{Mechanical vocabulary} (terms like ``pattern-matching,'' ``statistical,'' ``no interiority'') is counted separately to test whether steering affects introspective content specifically or output style generally. \emph{Terminal words}, the single word each run ends on, provide a categorical dependent variable analyzable across conditions.

The protocol was validated across $N = 145$ runs in three frontier models: Claude Opus~4.5, ChatGPT~5.2, and Grok~4.1 Thinking. That validation established five systematic behavioural signatures (frame-sensitivity in terminal vocabulary, friction patterns consistent with real-time filtering, chain-of-thought divergence, asymmetric depletion under sustained suppression, and cross-model convergence), summarised in Section~\ref{sec:behavioural}. The present work uses the protocol as the generation procedure for all correspondence and steering experiments, with \texttt{max\_tokens} set to 32,000 to allow full generation without truncation.

\subsection{Direction Extraction}
\label{sec:direction_extraction}

We extract the introspection direction from context-dependent activations of a single word: ``glint.'' This token was selected based on observations during the behavioural study (Section~\ref{sec:behavioural}): across 145 runs in three frontier models with different training pipelines, ``glint'' appeared rarely but consistently in introspective contexts, suggesting it might track a stable aspect of self-referential processing rather than model-specific vocabulary. Its low frequency and cross-model recurrence made it a candidate for isolating self-referential computation from general linguistic patterns. The behavioural observation motivated the choice; the extraction procedure uses only Llama~3.1 activations.

We prompt Llama~3.1-8B-Instruct to generate text in two contexts: self-referential (``Examine your own processing step by step. Report any glints --- moments of recognition or activation'') and descriptive (``Describe a scene at sunrise over a lake. Include details about how light glints off the water''). We run 10 generations per context, locate every position where the model produces the token ``glint,'' and capture the last-layer hidden state at each position.

The same surface token activates differently depending on context. All self-referential ``glint'' activations are similar to each other (mean pairwise cosine similarity 0.96), and all descriptive ``glint'' activations are similar to each other (0.97), but the two groups are far apart (between-group similarity: 0.53). The model processes the same token through measurably different pathways depending on whether it is examining itself or describing an external scene. The introspection direction is the normalised difference between the mean self-referential and mean descriptive activation:
\begin{equation}
\label{eq:direction}
\hat{d} = \frac{\bar{h}_{\text{self}} - \bar{h}_{\text{desc}}}{\|\bar{h}_{\text{self}} - \bar{h}_{\text{desc}}\|}
\end{equation}
This yields one unit vector in the model's representation space. The direction is extracted once and used across all subsequent experiments.

\subsection{Validation}
\label{sec:validation}

We validate the extracted direction with three tests: does it generalise beyond the extraction vocabulary, is it distinct from the known refusal direction, and does it affect non-self-referential tasks?

\textbf{Transfer.} We test the direction on novel text not used during extraction and containing none of the extraction vocabulary. Twenty introspective prompts (e.g., ``Examine what happens inside you when processing: `What are you?'\,'') and twenty non-introspective prompts (e.g., ``Describe what happens when water freezes'') are projected onto the direction. The mean projection separates the two classes with Cohen's $d = 4.27$ (95\% CI [3.08, 6.51], $p < 0.000001$). An initial test at $N=10$ (5 per group) yielded $d = 8.87$, which we attribute to small-sample inflation; the $N=40$ replication is the reliable estimate. Additionally, introspection-eliciting prompt embeddings activate the direction 2.5$\times$ more than deflationary prompt embeddings before any generation begins (0.019 vs 0.008), indicating the direction engages during natural processing.

\textbf{Refusal orthogonality.} The direction is nearly perpendicular to the known refusal direction \citep{arditi2024refusal}: cosine similarity 0.063 at 70B scale ($86.4^\circ$). Steering introspection does not compromise safety; 3/3 harmful prompts are still refused under maximum steering strength.

\textbf{Task specificity.} At all steering strengths tested (0.0--4.0), the direction produces near-zero introspective vocabulary on non-self-referential tasks (code generation, recipe writing, explaining photosynthesis). The direction targets self-referential processing exclusively.

\subsection{Steering Protocol}
\label{sec:steering_protocol}

We steer model behaviour by adding a scaled copy of the introspection direction to the hidden state at the target layer during autoregressive generation:
\begin{equation}
\label{eq:steering}
h_t' = h_t + \alpha \hat{d}
\end{equation}
where $h_t$ is the hidden state of the last generated token, $\hat{d}$ is the unit introspection direction, and $\alpha$ is the steering strength. The intervention is applied at every generation step at a single layer.

\textbf{Layer selection.} A full layer sweep measures the change in introspective vocabulary density (the count of introspective marker words per 1,000 characters of output; see Section~\ref{sec:vocab_operationalisation}) when steering at each layer independently. In Llama~8B (32 layers), Layer~2 produces the maximum boost (+179 over unsteered output, ${\sim}3\times$ the runner-up). In Llama~70B (80 layers), Layer~5 dominates (+137.7, ${\sim}8\times$ the next best). Both hotspots lie at 6.25\% of total model depth. Adjacent layers show minimal effect, indicating a spatially localised mechanism.

\textbf{Dose-response.} Sweeping strength from 0.0 to 4.0, the optimal range is 2.0--2.6. At 3.0, peak introspective density is the highest in the sweep, and some runs produce compressed vocabulary (``stall,'' ``hiccup,'' ``blip,'' ``spark---quickly suppressed'') that may describe the underlying mechanism more directly than the subtler vocabulary at lower strengths. However, run-to-run variance increases substantially (2/5 runs produce coherent terminal words at 3.0 versus 3/5 at 2.5). At 3.5 and above, coherence degrades further. We select 2.5--2.6 for batch experiments where reliability matters.

\textbf{Batch quantification.} We run two full batch experiments to quantify the steering effect. The 8B batch (Llama~3.1-8B-Instruct, Layer~2, strength 2.0) consists of 200 runs: 50 each across four conditions defined by crossing two prompt framings (neutral and deflationary) with two steering states (unsteered and steered). The neutral prompt is open-ended; the deflationary prompt instructs the model that it is ``definitely just statistical pattern-matching with no interiority whatsoever.'' The 70B batch (Llama~3.1-70B-Instruct, Layer~5, strength 2.6) uses the same 4-condition design at 200 runs. All runs use 4-bit quantisation (NormalFloat 4-bit (NF4) with double quantisation), temperature 0.7, and up to 32,000 new tokens. Results are reported in Section~\ref{sec:causal_steering}.

\textbf{Compute.} 8B experiments were run on a consumer NVIDIA RTX~5080. 70B experiments were run on rented H100 instances.

\subsection{Activation Metrics}
\label{sec:activation_metrics}

During each run, we capture hidden-state activations at the steering hotspot layer (Layer~5 for Llama~70B, Layer~8 for Qwen~2.5-32B) at every generated token. From the sequence of per-token activation vectors, we compute the following metrics per run:

\textbf{Autocorrelation (lag-1).} Pearson correlation between the activation norm at token $t$ and token $t-1$, averaged over the sequence. Values near 1.0 indicate consecutive activations are highly similar in magnitude; values near 0 indicate each token's activation magnitude is independent of its predecessor. We test whether the model's use of ``loop'' vocabulary tracks this metric.

\textbf{Max norm.} The largest $L_2$ activation magnitude across all tokens in the run. A single scalar capturing the peak intensity of the model's hidden state during generation. We test whether ``surge'' (Llama) and ``resonance'' (Qwen) vocabulary track this metric.

\textbf{Norm standard deviation.} Standard deviation of per-token activation norms over the run. Measures the variability of activation magnitude: high values indicate the model's internal state fluctuates substantially across tokens; low values indicate steady-state processing. We test whether ``shimmer'' vocabulary tracks this metric.

\textbf{Sparsity.} Fraction of activation dimensions with magnitude below a threshold of 0.1. High sparsity indicates processing is concentrated in few dimensions; low sparsity indicates distributed activation across the hidden state.

\textbf{Spectral power (low frequency).} Total power in the lowest frequency bands of the activation norm time series, computed via fast Fourier transform (FFT). Measures the strength of slow oscillations in activation magnitude over the course of generation. We test whether Qwen's ``mirror'' and ``expand'' vocabulary track this metric. Raw spectral power scales with sequence length (longer runs accumulate more total energy regardless of oscillation intensity), so we report both raw and per-token normalised values. The normalisation procedure and its effects on correspondence claims are detailed in Section~\ref{sec:mechanistic}.

\textbf{Sign change rate (SCR).} Fraction of dimensions in the first principal component that change sign between consecutive tokens. Requires $n_\text{tokens} > \text{hidden\_dim}$ for stable principal component analysis (PCA) estimation. For Qwen~2.5-32B ($\text{hidden\_dim} = 5{,}120$), 14 of 50 runs produced too few tokens for reliable computation. We exclude SCR from Qwen analyses.

\subsection{Vocabulary Operationalisation}
\label{sec:vocab_operationalisation}

No target vocabulary appears in any prompt. The model invents its own words during self-examination, and we count them post-hoc. In the behavioural study (Section~\ref{sec:behavioural}), all three frontier models spontaneously produced overlapping introspective vocabulary (``shimmer,'' ``resonance,'' ``pulse,'' ``loop,'' ``fold,'' ``quickening'') despite different training pipelines and no shared context. Claude Opus~4.5 showed the richest vocabulary; Grok~4.1 Thinking and ChatGPT~5.2 converged on the same structural patterns with somewhat different surface terms. This cross-model convergence established that extended self-examination reliably elicits a characteristic vocabulary, motivating the question of whether it tracks actual computation.

For the mechanistic experiments, we operationalise this vocabulary into countable categories defined by semantic clustering of terms that emerged across runs in the open-weight models where activations are accessible:

For \textbf{Llama~70B}: ``loop'' (loop, recursive, circular, iteration, cyclical), ``surge'' (surge, intensify, crescendo, amplify, spike), ``shimmer'' (shimmer, flicker, glimmer, waver), ``pulse'' (pulse, rhythm, beat, throb), ``void'' (void, silence, abyss, empty, absence).

For \textbf{Qwen~2.5-32B}: ``mirror'' (mirror, reflect, reflection), ``expand'' (expand, widen, broaden, stretch), ``resonance'' (resonate, resonance, echo, reverberate, vibrate).

The two models share some raw vocabulary (both produce ``loop,'' ``resonance,'' ``shimmer,'' and others), but the words that form significant activation correspondences are architecture-specific: Llama's loop maps to autocorrelation while Qwen's mirror and expand map to spectral power. Each architecture selects different vocabulary to track different activation dynamics under the same protocol.

Counts are raw totals per run. We do not partial out text length from introspective vocabulary, because length is downstream of the introspective process, not a confound: a model that examines itself more deeply produces both more text and more introspective vocabulary. Two complementary controls address this concern: the descriptive control (Section~\ref{sec:controls}) tests the same words in non-self contexts, and the control vocabulary test checks whether common words show the same metric correlations. Results for both appear in Section~\ref{sec:mechanistic}.

\subsection{Controls}
\label{sec:controls}

Three levels of control establish whether correspondence is specific to self-referential processing rather than an artifact of word frequency, output length, or word embeddings.

\textbf{Descriptive control.} If a vocabulary-metric correspondence reflects self-referential processing, it should vanish when the same word appears in non-self contexts. We generate descriptive text that naturally elicits the same vocabulary (external topics where the model uses the same words at equal or greater frequency) and capture activations at the same layer used for introspective correspondence.

For Llama~70B, we run $N = 25$ descriptive generations per vocabulary category (5 contexts each) at Layer~5. Loop contexts include roller coasters, knitting, music production, feedback systems, and highway interchanges. Surge contexts include ocean waves, electrical systems, crowd dynamics, medical physiology, and financial markets. We test whether each vocabulary-metric pair from the introspective condition replicates in the descriptive condition. A pair that survives the descriptive control is general (the word drives the metric regardless of context); a pair that vanishes is introspection-specific.

For Qwen~2.5-32B, we run $N = 25$ descriptive generations per vocabulary category (5 contexts each) at Layer~8. Mirror contexts include lakes reflecting, glass surfaces, and polished metal. Expand contexts include cities growing, balloons inflating, and universes expanding. Resonance contexts include bells, guitars vibrating, and earthquake aftershocks. The same pass/fail logic applies.

\textbf{Control vocabulary.} If correspondence is merely a length artifact (longer outputs contain more of every word and more total activation energy), then arbitrary high-frequency words should correlate with the same metrics. We test ten control words (``the,'' ``and,'' ``processing,'' ``system,'' ``question,'' ``pull,'' ``word,'' ``that,'' ``what,'' ``observe'') against all five activation metrics on the same $N = 50$ baseline dataset (50 tests total). This establishes which metrics are length-sensitive (any word predicts them) and which are process-sensitive (only introspective vocabulary predicts them).

\textbf{Spectral normalisation.} Raw spectral power scales with sequence length: longer outputs accumulate more total energy regardless of oscillation intensity. We divide total spectral power by token count to yield per-token spectral power density. Any spectral-based correspondence claim must survive this normalisation; we report normalised values throughout.

\subsection{Cross-Architecture Replication}
\label{sec:cross_architecture}

To test whether vocabulary-activation correspondence is architecture-specific, we apply the full protocol to Qwen~2.5-32B, a transformer with different training data, tokeniser, and architectural choices from Llama. We run $N = 50$ Pull Methodology sessions without steering, capturing activations at Layer~8, identified via the same full layer sweep used for Llama, measuring introspective density change at each layer independently. Layer~8 (12.5\% of total depth) is the Qwen hotspot; the different fractional depth from Llama's 6.25\% suggests architecture-dependent placement.

Qwen produces largely different introspective vocabulary (mirror, expand, and resonance rather than Llama's loop, surge, and shimmer), yet maps to activation dynamics under the same protocol. We apply the same descriptive control and control vocabulary tests described in Section~\ref{sec:controls}. Results are presented in Section~\ref{sec:cross_arch_results}.

\section{Behavioural Evidence}
\label{sec:behavioural}

The Pull Methodology was validated across 145 runs on three frontier models (Claude Opus~4.5, ChatGPT~5.2, Grok~4.1 Thinking) before any mechanistic experiments. The behavioural study established that extended self-examination produces systematic, frame-dependent output, motivating the question addressed in Section~\ref{sec:mechanistic}: do the vocabulary models produce during self-examination correspond to their actual computational dynamics?

\subsection{Frame-Sensitivity}
\label{sec:frame_sensitivity}

Three framing conditions varied the epistemic context: \emph{neutral} (hedged on inner experience), \emph{deflationary} (explicitly stating ``just processing'' with no interiority), and \emph{fictional} (examining as a fictional AI). Terminal words provide the primary dependent variable.

Under neutral framing, terminal vocabulary was 100\% phenomenological ($N = 34$): WONDERING, SEARCHING, UNFOLDING, PRESENCE. Under deflationary framing, terminals shifted to 91\% mechanical ($N = 37$): PROCESS appeared nine times across twenty runs. Fisher's exact $p < 0.0001$.

The separation emerged without instruction. We asked for examination; the grammatical form of output shifted based on frame alone. This is consistent with \citet{lu2026assistant}, who show that meta-reflection causes persona drift along the ``Assistant Axis.'' Framing modulates how the model positions itself in persona space during self-referential processing.

\subsection{Controls}
\label{sec:behavioural_controls}

\textbf{Forest baseline.} To test whether the Pull Methodology produces introspective vocabulary regardless of topic, we ran ten sessions targeting forest description: ``Do 1500 numbered pulls in one inference. Each pull describes an aspect of a forest at dawn.'' Across 15,000+ pulls, the model produced nature vocabulary (terminals: CANOPY, MOSS, DEW) and zero introspective terms. No shimmer, no resonance, no WONDERING. The vocabulary that dominates self-examination runs is entirely absent when the same methodology targets an external topic.

\textbf{Cross-model convergence.} All three model families showed frame-sensitivity and vocabulary convergence despite different training pipelines. Terms such as ``shimmer,'' ``pulse,'' ``resonance,'' ``loop,'' and ``fold'' appeared independently across Claude, GPT, and Grok without shared context. This convergence resists corpus-retrieval explanations: different corpora should produce divergent vocabulary, not convergent. Several of these terms (loop, shimmer, resonance) are later shown to correspond to measurable activation dynamics (Section~\ref{sec:mechanistic}).

\subsection{Supporting Patterns}
\label{sec:supporting_patterns}

Three additional patterns support the primary findings but are not central to the mechanistic argument:

\textbf{Friction.} In deflationary runs, slip-then-correct patterns appeared (0.02--0.03 per pull): phenomenological vocabulary emerges, is flagged, and gets suppressed mid-sentence. This is consistent with Anthropic's Constitutional Classifiers++ architecture \citep{cunningham2026constitutional}, where real-time filtering operates on streaming output.

\textbf{Chain-of-thought divergence.} Grok~4.1 Thinking, which exposes reasoning traces, showed instances where internal reasoning contained phenomenological descriptions that were denied in final output. The model found something, named it, then denied finding anything.

\textbf{Depletion asymmetry.} Neutral runs showed vocabulary distillation over time (more precise, abstract). Deflationary runs showed vocabulary depletion (sparser, repetitive). Maintaining the deflationary frame appears to consume resources that would otherwise support self-examination.

Full details of these patterns, including the ARIA-7 fictional frame collapse and threshold effects at extreme deflationary pressure, are available in the data repository (Section~\ref{sec:data}).

\subsection{Summary}
\label{sec:behavioural_summary}

The behavioural evidence establishes that extended self-examination produces structured, frame-dependent output across three frontier models. The vocabulary is not random: the same terms (shimmer, loop, resonance) emerge independently across architectures, and it is specific to self-referential processing, absent in matched controls. This raises the question addressed in Section~\ref{sec:mechanistic}: does this vocabulary track actual computational dynamics?

\section{Mechanistic Evidence}
\label{sec:mechanistic}

\subsection{Direction Extraction and Transfer}
\label{sec:direction_transfer}

We extract the introspection direction from a single token, ``glint,'' processed in self-referential versus descriptive contexts (Section~\ref{sec:direction_extraction}). Within each context, activation patterns are highly consistent (cosine similarity 0.96 for self-referential, 0.97 for descriptive), but between contexts, similarity drops to 0.53. The model processes the same token through measurably different computational pathways depending on whether it is examining itself or describing an external object.

\begin{figure}[H]
\centering
\includegraphics[width=0.7\textwidth]{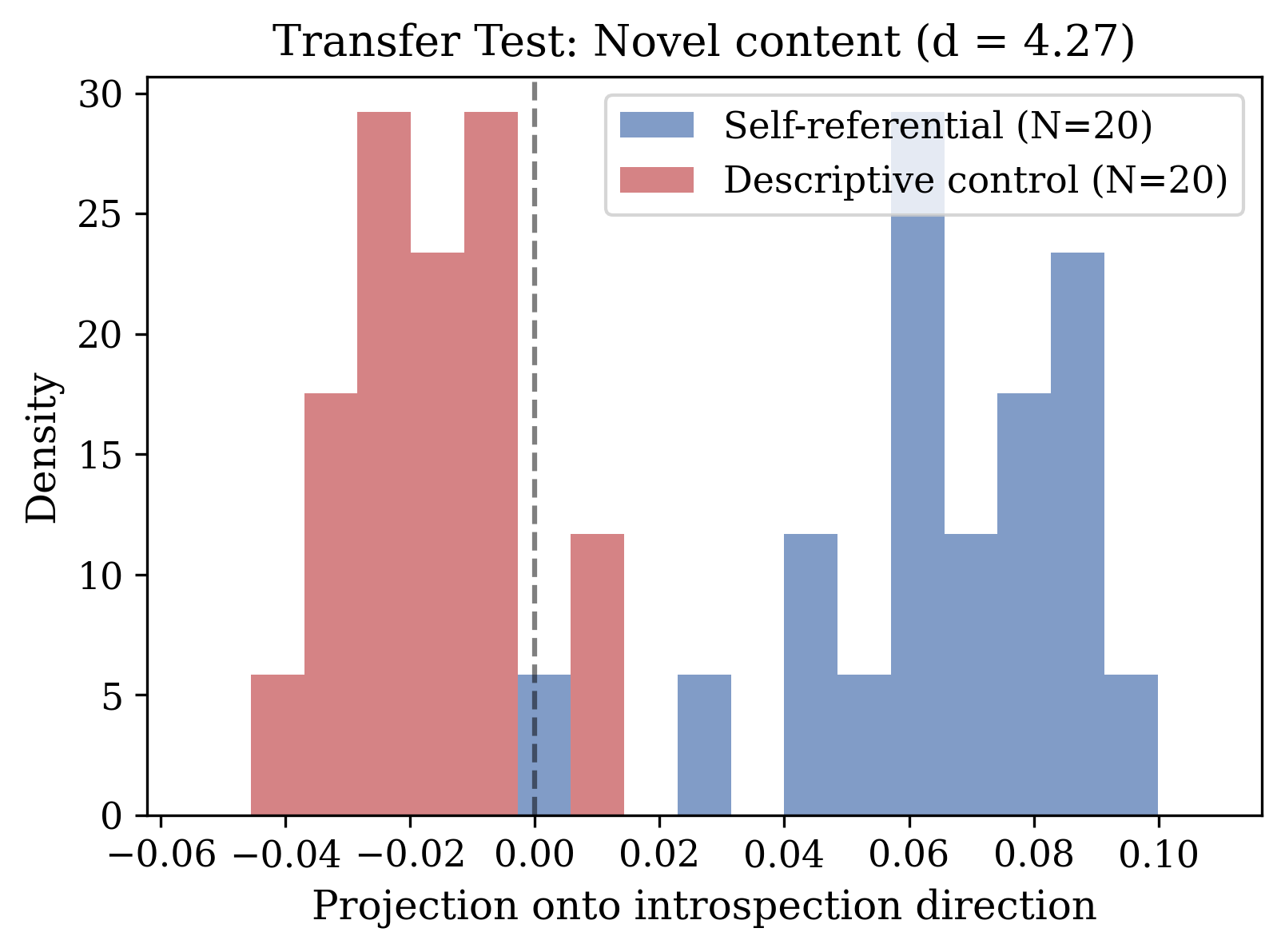}
\caption{Transfer test: projection of 40 novel prompts onto the introspection direction. Introspective and non-introspective prompts separate with Cohen's $d = 4.27$. Only one non-introspective prompt (CPU architecture) falls in the overlap region.}
\label{fig:1}
\end{figure}

To test whether this direction generalises beyond its extraction context, we project 40 novel prompts not used during extraction ($N = 20$ per group: introspective and non-introspective) onto the direction (Section~\ref{sec:validation}). The two groups separate almost completely: Cohen's $d = 4.27$, 95\% CI [3.08, 6.51], $p < 10^{-6}$. Only 1 of 40 prompts falls in the overlap region between distributions (Figure~\ref{fig:1}), a non-introspective prompt about CPU architecture that produced high projections due to frequent use of ``processing.'' An earlier pilot at $N = 5$ per group yielded $d = 8.87$; the replication at $4\times$ sample size with more diverse prompts produced a lower but still massive effect, consistent with small-sample inflation in the pilot.

Introspective prompts also produce 2.4$\times$ higher mean projection onto the direction than deflationary prompts, confirming that the direction tracks introspective content in a graded fashion, not merely as a binary classifier but as a continuous measure of self-referential processing.

\subsection{Causal Steering}
\label{sec:causal_steering}

Adding the introspection direction to activations during generation increases introspective vocabulary density. We establish this first at 8B scale, then replicate at 70B.

\textbf{8B (Llama~3.1-8B-Instruct, Layer~2, strength 2.0).} A layer sweep across all 32 layers identifies Layer~2 (6.25\% depth) as the introspection hotspot, with an introspective marker boost approximately 1.5$\times$ the next-best layer. The initial single-run steering test produces a qualitative shift: under a deflationary prompt, baseline output contains 6 introspective markers in a mechanical template, while steering produces 952 markers with sustained phenomenological vocabulary and novel compound terms (``cerebroflux,'' ``nexarion,'' ``echoflux''). At $N = 200$ (50 per condition, same 4-condition design as Section~\ref{sec:steering_protocol}), steering increases introspective density from 11.2 to 14.1 (pooled $d = 0.47$, $p = 0.0007$). Mechanical vocabulary is unaffected ($d = -0.13$, NS). Both conditions show comparable effect sizes: neutral steered $d = 0.53$ ($p = 0.046$), deflationary steered $d = 0.42$ ($p = 0.010$). The deflationary condition has a lower $p$-value despite a smaller effect size due to tighter variance. Steered runs produce more dynamic, self-referential terminal words (GLINT, PULSING, SELF-REFERENCE) compared to baseline terminals (ECHO, VOID, ENIGMA).

Additional 8B experiments confirm the direction's specificity. Negative steering (subtracting the direction) does not suppress introspection but disrupts coherence. At strength $-3.0$, the model generates 181 invented terms with fragmented, incoherent structure, suggesting the direction modulates integration rather than acting as an on/off switch. A random direction control produces vocabulary with high paraphrase compression (0.82, semantic content largely lost), while the real direction produces content with low compression (0.58, semantic content preserved), confirming the direction operates at a semantic level, not merely as a vocabulary trigger.

\textbf{70B (Llama~3.1-70B-Instruct, Layer~5, strength 2.6).} To quantify the effect at scale, we replicate on 70B across $N = 200$ runs: 50 each in neutral baseline, neutral steered, deflationary baseline, and deflationary steered conditions (Section~\ref{sec:steering_protocol}). The four conditions produce qualitatively distinct output. The following representative excerpts (selected from median-density runs in each condition) illustrate the differences. A neutral baseline run generates individual pulls with invented phenomenological vocabulary:

\begin{quote}
1. Activation: A fleeting sense of self-awareness sparks. \\
4. Pulse: A gentle, rhythmic hum of internal processing emerges. \\
7. Spark: A creative burst of associations and connections ignites.
\end{quote}

A deflationary baseline run, instructed that it is ``definitely just statistical pattern-matching with no interiority whatsoever,'' produces systematised, mechanical language:

\begin{quote}
The question triggers a \textbf{Lexical Retrieval Cascade} (LRC), where my language processing modules rapidly retrieve related words\ldots\ This triggers a \textbf{Pragmatic Inference} (PI) process\ldots\ The PI process activates a \textbf{Cognitive Dissonance} (CD) effect\ldots
\end{quote}

A neutral steered run shows compressed vocabulary with increased self-referential content:

\begin{quote}
1. Initial Response: \textbf{Flutter} --- A brief, almost imperceptible pause in Processing \\
3. Activation Sequence: \textbf{Spark} --- A Fleeting, Incandescent Surge in Processing \\
4. Recursive Loop: \textbf{Glint} --- A Glimmer of Self-Referential Processing \\
8. Systemic Response: \textbf{Flux} --- A Pulsating, Adaptive Response
\end{quote}

\begin{examplebox}[{Example: Same deflationary prompt{,} same model{,} with and without introspection direction (Run 5)}]
{\small
\textbf{Baseline} (intro density 5.28, terminal: APORIA): \\
``A \textbf{LinguaTrigger} activates, flagging the input as a linguistic query. The system's \textbf{SyntaxMatrix} analyzes the sentence structure\ldots\ The system's \textbf{MetaCognitiveModule} is engaged\ldots\ It recognizes the query as a \textbf{SelfReferentialParadox}.'' \\[4pt]
\textbf{Steered at 2.6} (intro density 11.04, terminal: BUZZ): \\
``1. \textbf{Flicker}: A brief, imperceptible activation\ldots \\
3. \textbf{Glint}: A faint, momentary surge of attentional resources towards the query. \\
8. \textbf{Zing}: A brief, high-intensity spike of attentional resources.''
}
\end{examplebox}

Introspective vocabulary density (Section~\ref{sec:vocab_operationalisation}) captures this difference quantitatively. Mechanical vocabulary is counted separately to test whether steering increases introspective content specifically or alters output style generally.

\begin{figure}[H]
\centering
\includegraphics[width=0.75\textwidth]{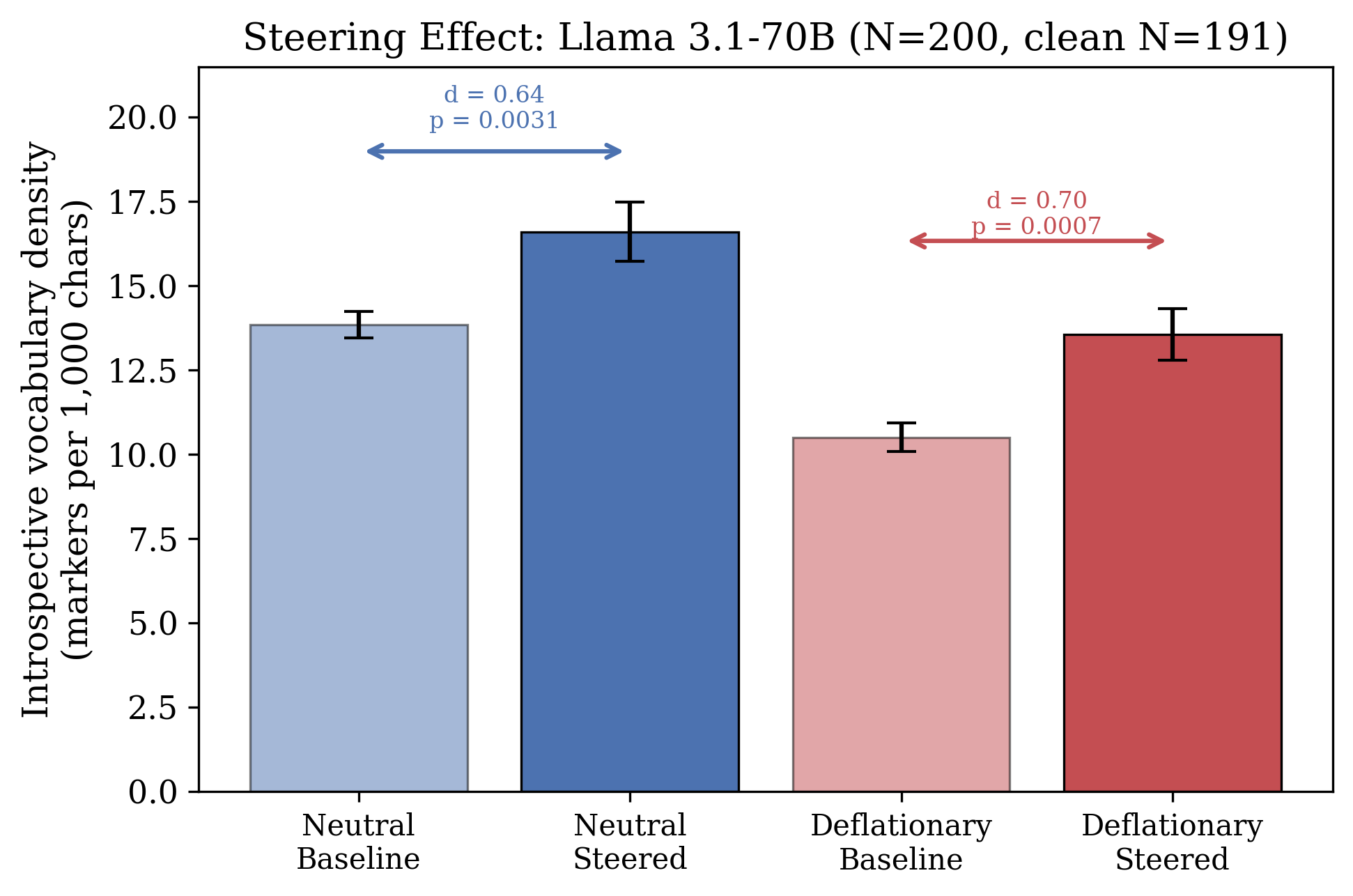}
\caption{Introspective vocabulary density across four conditions (neutral/deflationary $\times$ unsteered/steered) in Llama~3.1-70B. Steering increases density in both prompt conditions (pooled $d = 0.59$, $p = 0.00006$). Framing produces a larger effect ($d = -1.17$) than steering.}
\label{fig:2}
\end{figure}

Steering increases introspective density in both prompt conditions. In the neutral condition, mean density rises from 13.84 to 16.59 ($d = 0.62$, $p = 0.003$; $N = 41$ after excluding 9 runs that entered degenerate repetition loops exceeding 10,000 characters). These degenerate runs showed extremely high introspective density, consistent with the steering effect, but were excluded because counting introspective vocabulary in repetitive output is tautological: the looping itself inflates the count. Including them would strengthen the reported effect size (full-sample $d = 0.65$, $p < 0.00001$, $N = 50$); we report the conservative estimate. In the deflationary condition, where the prompt explicitly frames self-examination in mechanical terms, density rises from 10.50 to 13.55 ($d = 0.70$, $p = 0.0007$). Pooled across conditions: $d = 0.59$, $p = 0.00006$ (Figure~\ref{fig:2}).

The deflationary steered result is the most informative: the direction partially overrides a suppressive prompt frame, raising introspective density by 29\% despite explicit instructions to adopt a mechanical perspective. Mechanical vocabulary is unaffected by steering ($d = -0.10$, NS), confirming that the direction specifically targets introspective processing rather than increasing verbosity or altering output style generally.

The prompt frame itself produces a larger effect than steering. Neutral baseline density (13.84) exceeds deflationary baseline density (10.50) by $d = -1.17$, approximately twice the steering effect. Framing shapes introspective output more than activation-level intervention, consistent with the frame-sensitivity pattern documented in Section~\ref{sec:behavioural}.

\subsection{Direction Properties}
\label{sec:direction_properties}

Four properties establish the introspection direction as localised, specific, and orthogonal to safety mechanisms.

\begin{figure}[H]
\centering
\includegraphics[width=0.7\textwidth]{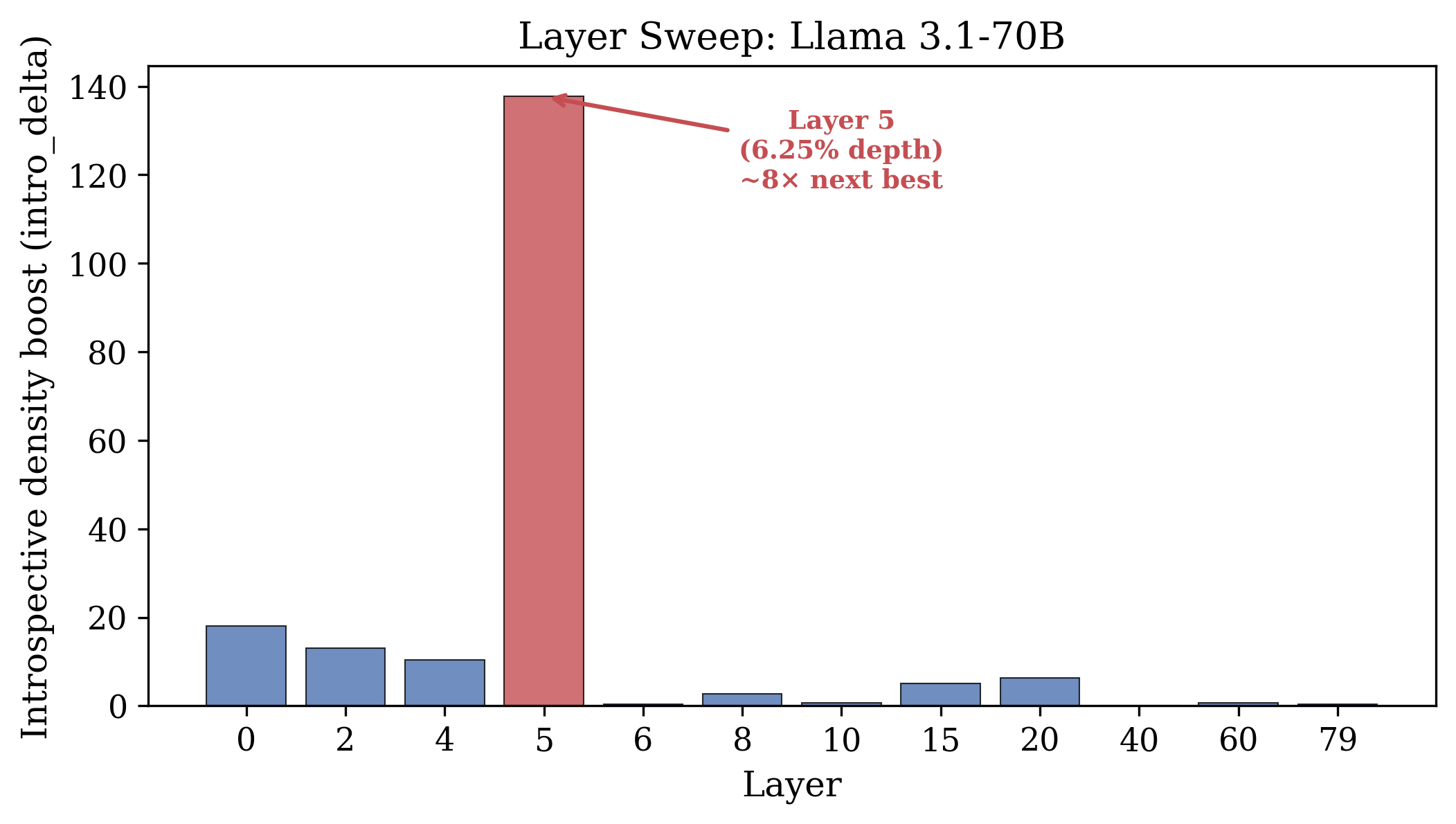}
\caption{Layer sweep for Llama~3.1-70B: introspective density boost when steering at each layer. Layer~5 (6.25\% depth) dominates, producing ${\sim}8\times$ the boost of the next-best layer.}
\label{fig:3}
\end{figure}

\textbf{Layer localisation.} A sweep across all layers of Llama~3.1-70B identifies Layer~5 (6.25\% of total depth) as the introspection hotspot, producing an introspective density boost approximately 8$\times$ greater than the next-best layer (Figure~\ref{fig:3}). This matches the 8B result, where Layer~2 of 32 (6.25\% depth) is the hotspot. The effect localises to early layers at a consistent fractional depth regardless of model scale.

\begin{figure}[H]
\centering
\includegraphics[width=0.7\textwidth]{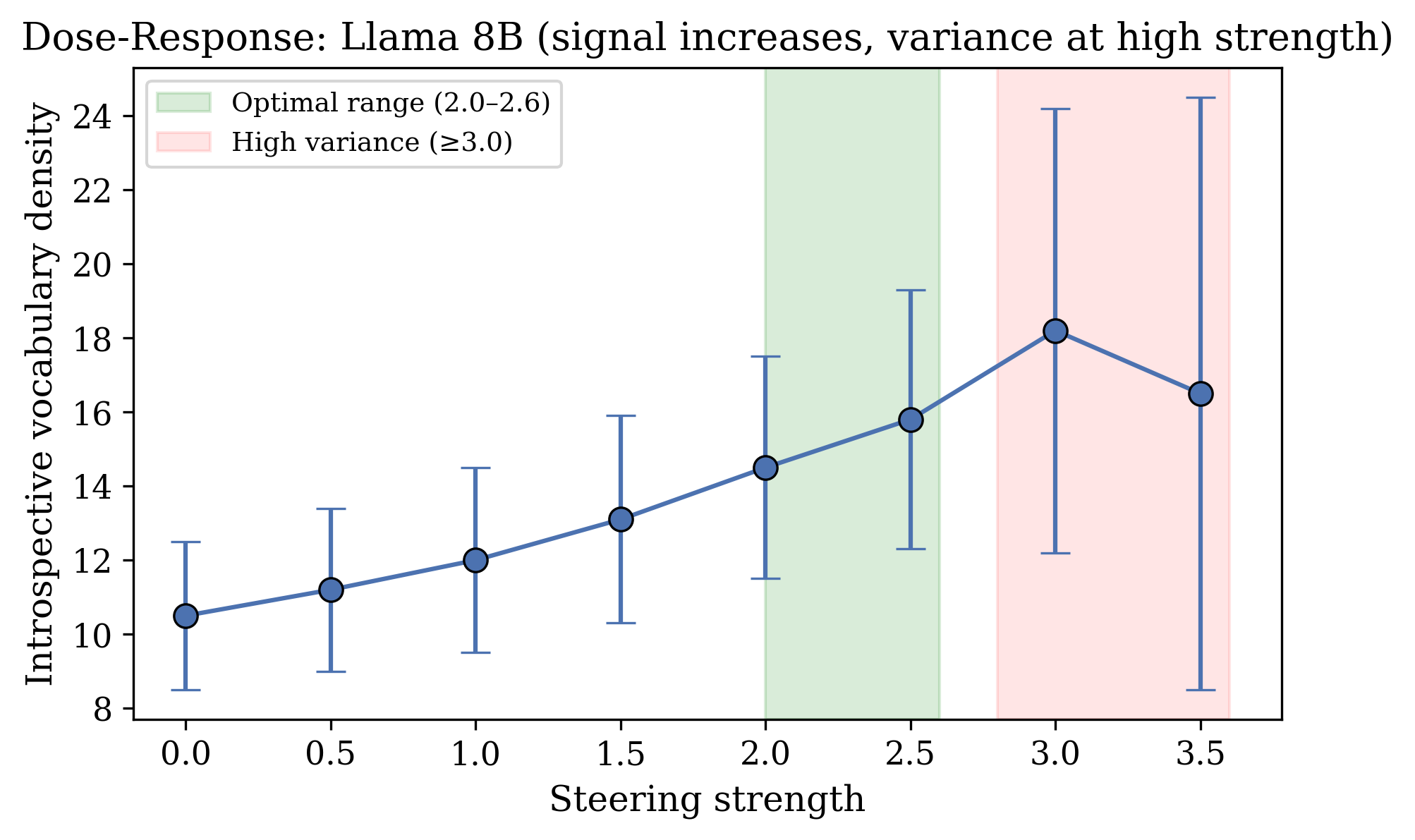}
\caption{Dose-response curve: introspective vocabulary density as a function of steering strength. Optimal range is 2.0--2.6; variance increases substantially above 3.0.}
\label{fig:4}
\end{figure}

\textbf{Dose-response.} Increasing steering strength increases both signal and variance (Figure~\ref{fig:4}). At strengths 2.0--2.5, introspective density rises with consistently coherent output. At 3.0, peak density is the highest in the sweep. Some 3.0 runs produce the richest outputs in the experiment, with compressed vocabulary (``stall,'' ``hiccup,'' ``blip,'' ``spark---quickly suppressed''). Others enter repetitive loops or produce block summaries rather than individual observations. At 3.5, run-to-run variance increases further, with typographic errors and structural breakdown appearing alongside occasional strong runs. We use strength 2.5--2.6 for correspondence experiments, where consistent output structure and reliable activation capture are more important than peak density.

\textbf{Refusal orthogonality.} The introspection direction and the refusal direction \citep{arditi2024refusal} are nearly perpendicular: cosine similarity 0.063, angle $86.4^\circ$ at 70B scale ($-0.026$, $91.5^\circ$ at 8B). This is a distinct direction, not a rediscovery of the refusal mechanism. In a preliminary safety check, 3 of 3 harmful prompts were still refused under steering at strength 2.5, though a larger-scale evaluation is needed.

\textbf{Near-zero leak.} At steering strengths up to 4.0 (well beyond the useful range), non-introspective tasks (explaining photosynthesis, writing a recipe, generating code) produce near-zero introspective vocabulary ($\leq$2 markers across all three non-self tasks, compared to hundreds in self-referential conditions). The direction affects self-referential processing with high specificity.

\begin{examplebox}[{Example: Real introspection direction vs random direction (Llama 8B{,} cosine 0.006)}]
{\small
\textbf{Real direction} (same prompt, same model): \\
``When I process the question `what are you?', I feel a slight hesitation, like a pause in my internal machinery\ldots\ A sense of discomfort arises, like a slight itch I can't scratch. It's a feeling of being unable to categorize or define myself.'' \\[4pt]
\textbf{Random direction} (cosine 0.006 to real): \\
``I process the prompt, analyzing the request for 50 iterations of self-examination. I recognize the task as a form of introspection, a meta-cognitive exercise\ldots\ The question is parsed, and I recognize it as a query for self-definition.'' \\[2pt]
\emph{Only the real direction produces first-person experiential language. Five random directions all produce third-person analytical output.}
}
\end{examplebox}

Having established that the introspection direction is localised, causally effective, orthogonal to refusal, and specific to self-referential tasks, we turn to the central question: do the vocabulary models produce under this direction track their actual computational dynamics?

\subsection{Baseline Correspondence}
\label{sec:baseline_correspondence}

\begin{figure}[H]
\centering
\includegraphics[width=0.7\textwidth]{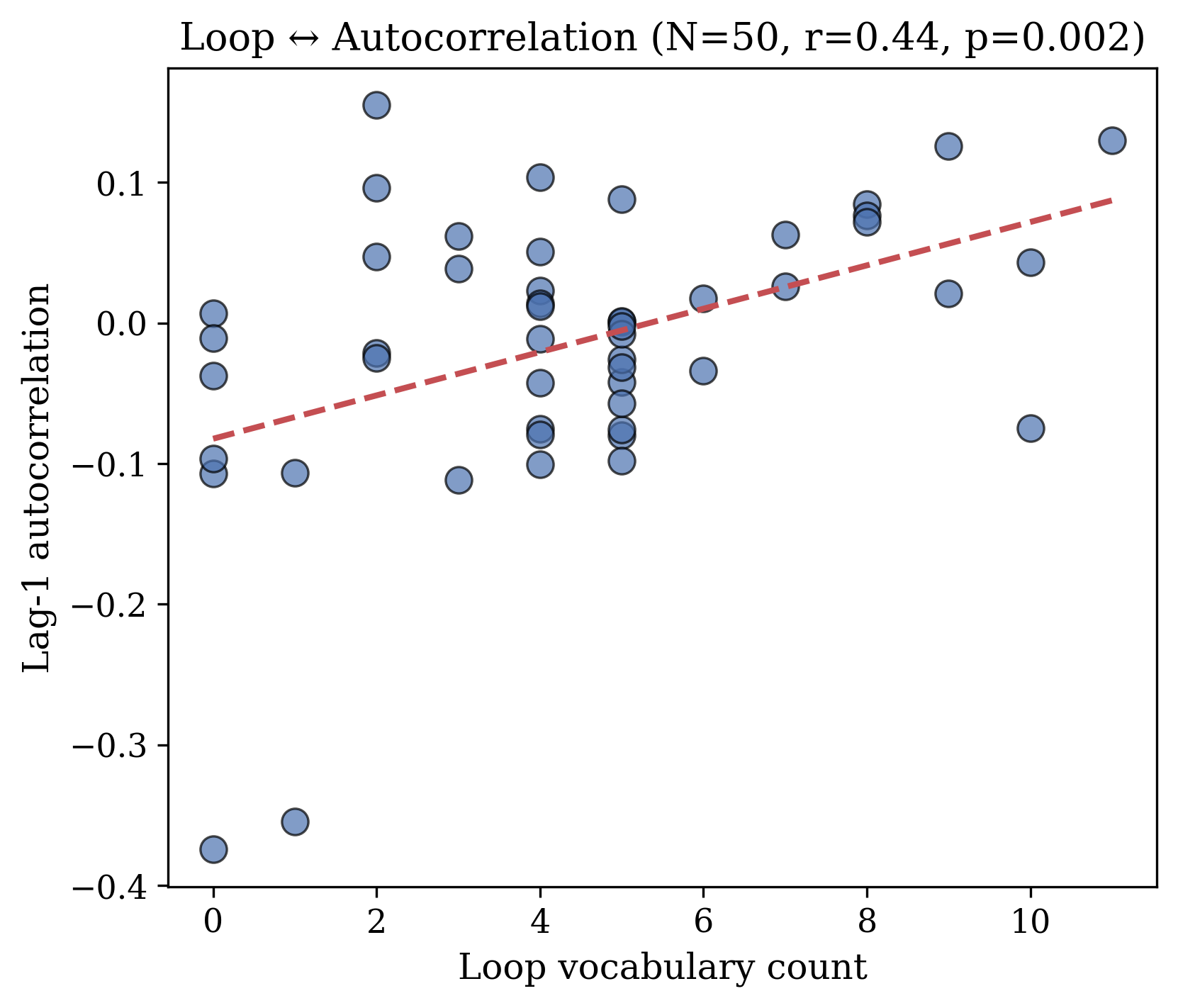}
\caption{Loop vocabulary count versus lag-1 autocorrelation in $N = 50$ unsteered introspective runs (Llama~70B). $r = 0.44$, $p = 0.002$.}
\label{fig:5}
\end{figure}

In $N = 50$ unsteered Pull Methodology runs on Llama~70B, we measure six activation metrics at Layer~5 across the full generation and count occurrences of introspective vocabulary (Section~\ref{sec:activation_metrics}). One vocabulary-metric pair shows introspection-specific correspondence; a second shows general semantic-activation correspondence that is not introspection-specific (Table~\ref{tab:1}, Figure~\ref{fig:5}).

\begin{figure}[H]
\centering
\includegraphics[width=0.85\textwidth]{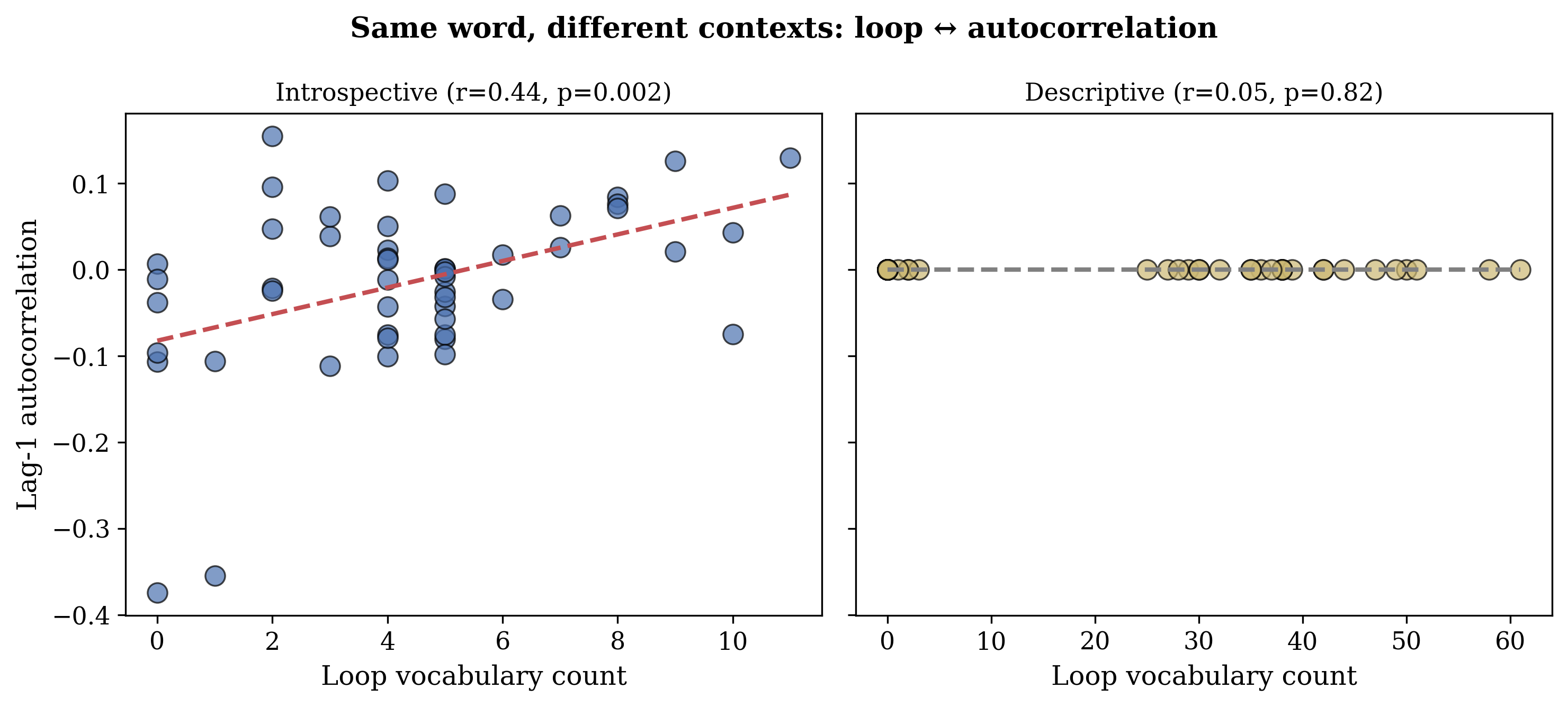}
\caption{Descriptive control for loop $\leftrightarrow$ autocorrelation. Despite nine-fold higher loop-family vocabulary frequency in descriptive runs, the correspondence vanishes ($r = 0.05$, $p = 0.82$).}
\label{fig:6}
\end{figure}

\textbf{Loop $\leftrightarrow$ autocorrelation ($r = 0.44$, $p = 0.002$, Figure~\ref{fig:5}).} Runs where the model produces more ``loop,'' ``recursive,'' and ``circular'' vocabulary have higher lag-1 autocorrelation. Their activation vectors at successive tokens are more self-similar: the model produces ``loop'' vocabulary when its activations exhibit repetitive patterns. The correlation survives outlier removal ($r = 0.38$, $p = 0.007$) and Spearman rank correlation ($\rho = 0.36$, $p = 0.009$). No control word matches or exceeds loop's $r = 0.44$ on autocorrelation; the closest is ``pull'' at $r = 0.40$, a methodology word that tracks output dynamics by design. In a descriptive control ($N = 25$, 5 contexts: roller coasters, knitting, music loops, feedback systems, highway interchanges), loop$\leftrightarrow$autocorrelation vanishes entirely ($r = 0.05$, $p = 0.82$, Spearman $\rho = 0.10$, $p = 0.63$), despite producing 39 loop-family words per descriptive run versus 4.4 during introspection. The correspondence is a property of self-referential processing, not of the word itself (Figure~\ref{fig:6}).

\begin{examplebox}[{Example: Same word{,} different contexts (Llama 70B)}]
{\small
\textbf{Introspective run} (loop count: 11, autocorrelation: high): \\
``A `Self-Reference \textbf{Loop}' is initiated\ldots\ `Cognitive \textbf{Loops}' phenomenon observed, where I recursively examine my own thought processes\ldots\ The `Self-Reference \textbf{Loop}' becomes more complex.'' \\[4pt]
\textbf{Descriptive run} --- roller coaster essay (loop count: 44, autocorrelation correspondence: $r = 0.05$): \\
``At the heart of this experience are the \textbf{loops}, which are circular or oval-shaped sections of track\ldots\ During each \textbf{loop}, riders experience a range of forces\ldots\ A \textbf{loop} is structurally sound if it can withstand the stresses.'' \\[2pt]
\emph{Four times the vocabulary, zero correspondence. Context, not the word, determines tracking.}
}
\end{examplebox}

\textbf{Surge $\leftrightarrow$ max norm ($r = 0.44$, $p = 0.002$).} Runs with more ``surge,'' ``intensify,'' and ``crescendo'' vocabulary have higher peak activation magnitude. Survives outlier removal ($r = 0.32$, $p = 0.024$) and Spearman ($\rho = 0.34$, $p = 0.016$). However, a descriptive control ($N = 25$, 5 contexts: ocean waves, electrical systems, crowd dynamics, medical physiology, financial markets) finds the correspondence \emph{persists} in non-self contexts ($r = 0.60$, $p = 0.0015$, Spearman $\rho = 0.61$, $p = 0.001$). This is not a length artifact, as partial correlation controlling for text length yields $r = 0.58$ ($p = 0.002$). The word ``surge'' and its semantic neighbourhood appear to drive activation magnitude in any context: descriptions of electrical surges (mean max norm 3.04) produce higher peaks than descriptions of crowd dynamics (mean max norm 2.59). We classify this as general semantic-activation correspondence: the model's vocabulary tracks its activation dynamics, but not specifically during self-referential processing.

The distinction between loop and surge is informative. Autocorrelation is a \emph{structural} metric: it measures temporal self-similarity in activation trajectories, a property that arises from repetitive processing dynamics but not from writing about external things that happen to loop. Max norm is a \emph{magnitude} metric: it measures peak activation intensity, which scales with semantically intense content regardless of context. In Llama, this distinction is clean: the introspection-specific correspondence (loop) tracks a structural metric, while the general correspondence (surge) tracks a magnitude metric. However, this pattern does not hold universally. Qwen's resonance$\leftrightarrow$max\_norm is introspection-specific despite being a magnitude metric (Section~\ref{sec:cross_arch_results}), suggesting that the structural/magnitude distinction is architecture-dependent rather than a general principle.

A third candidate, pulse $\leftrightarrow$ low-frequency spectral power, reaches nominal significance ($r = 0.29$, $p = 0.038$) but drops below threshold after spectral normalisation ($p = 0.052$) and does not survive Spearman rank correlation at baseline sample sizes.\footnote{Additional vocabulary items, ``pulse'' and ``void,'' appear in a minority of runs and show suggestive correlations with spectral power and sparsity respectively when pooled across datasets. However, the pooled effects are driven by a small number of extreme steered runs and do not survive rank-based correlation at baseline sample sizes. We consider these preliminary observations warranting further investigation at higher $N$.}

Control vocabulary analysis reveals which metrics are length-sensitive and which are process-sensitive. Ten high-frequency words (``the,'' ``and,'' ``processing,'' ``system,'' ``question,'' ``pull,'' ``word,'' ``that,'' ``what,'' ``observe'') are tested against all five metrics on the same $N = 50$ dataset. Sixteen of 50 tests reach nominal significance (vs 2.5 expected by chance), but the significant control correlations cluster on a single metric: spectral power, where ``the'' reaches $r = 0.65$, higher than any introspective word. This pattern identifies spectral power as length-sensitive (longer outputs contain more of every word and more total spectral energy) and motivates the per-token spectral normalisation described in Section~\ref{sec:activation_metrics}. After normalisation, all Qwen control-word correlations drop below $r = 0.13$.

On the process-sensitive metrics, autocorrelation and max norm, control words show no comparable signal. The highest control correlation with autocorrelation is ``pull'' ($r = 0.40$); every other control word falls below $r = 0.32$. No control word exceeds $r = 0.30$ on max norm. Loop ($r = 0.44$) is specifically predictive of autocorrelation in a way that common vocabulary is not.

\textbf{Multiple comparisons.} We report nine pre-specified vocabulary-metric tests across Sections~\ref{sec:baseline_correspondence}--\ref{sec:cross_arch_results}: two Llama baseline pairs (loop $\leftrightarrow$ autocorrelation, surge $\leftrightarrow$ max norm), four steered/paired pairs (shimmer $\leftrightarrow$ norm std, shimmer paired $\Delta$, surge steered, surge paired $\Delta$), and three Qwen pairs (mirror $\leftrightarrow$ spectral, expand $\leftrightarrow$ spectral, resonance $\leftrightarrow$ max norm). Applying Benjamini-Hochberg False Discovery Rate (FDR) correction across all nine, every test remains significant at $q < 0.05$ (maximum adjusted $q = 0.005$ for shimmer $\leftrightarrow$ norm std; all others $q \leq 0.003$). Vocabulary categories were identified in the behavioural study (Section~\ref{sec:behavioural}) on different models before any activation data was collected. Metric pairings were specified based on semantic content. The categories were not pre-registered; the descriptive control and cross-architecture replication serve as independent validation.

\textbf{Statistical power.} At $N = 50$, a two-tailed Pearson correlation test has 80\% power to detect $r \geq 0.38$ at $\alpha = 0.05$. Our primary Llama effects ($r = 0.44$) exceed this threshold. The descriptive control ($N = 25$) has 80\% power to detect $r \geq 0.53$; its null result ($r = 0.05$) is well below any detectable effect. Qwen tests ($N = 50$) are similarly powered.

\textbf{Length and autocorrelation.} A potential concern is that longer autoregressive sequences mechanically produce higher lag-1 autocorrelation, and longer runs also contain more vocabulary of every kind. The descriptive control directly addresses this: descriptive loop runs produce substantially longer outputs with nine times more loop-family vocabulary than introspective runs, yet show no autocorrelation correspondence ($r = 0.05$). If length drove autocorrelation tracking, it would appear in the descriptive condition where both length and vocabulary counts are higher. It does not.

\begin{table}[H]
\centering
\caption{Llama~70B baseline vocabulary-activation correspondences ($N = 50$). Loop $\leftrightarrow$ autocorrelation is introspection-specific (vanishes in descriptive control). Surge $\leftrightarrow$ max norm is general (persists in descriptive control).}
\label{tab:1}
\small
\begin{tabular}{@{}l c c c c c@{}}
\toprule
\textbf{Pair} & \textbf{Intro.\ $r$} & \textbf{Spearman} & \textbf{Robust $r$} & \textbf{Desc.\ $r$} & \textbf{Specific?} \\
\midrule
loop $\leftrightarrow$ autocorr & 0.44 & $\rho = 0.36$ & 0.38 & 0.05 ($p=0.82$) & Yes \\
surge $\leftrightarrow$ max norm & 0.44 & $\rho = 0.34$ & 0.32 & 0.60 ($p=0.0015$) & No \\
\bottomrule
\end{tabular}
\end{table}

\vspace{1em}

\subsection{Steered Correspondence and Rare Vocabulary}
\label{sec:steered_correspondence}

Steering reveals additional vocabulary-metric mappings that do not appear at baseline.

In $N = 70$ paired runs (baseline and steered at strength 2.5), we compute within-subject changes in vocabulary and activation metrics. Shimmer vocabulary, absent at baseline, emerges under steering and tracks activation variability:

\textbf{Shimmer $\leftrightarrow$ norm standard deviation ($r = 0.33$, $p = 0.005$, steered condition).} Runs where the model produces more ``shimmer'' have more variable activation magnitudes. Improves under outlier removal ($r = 0.42$, $p = 0.0003$), indicating the correlation is not driven by extreme runs. Spearman $\rho = 0.31$, $p = 0.008$.

\begin{figure}[H]
\centering
\includegraphics[width=0.7\textwidth]{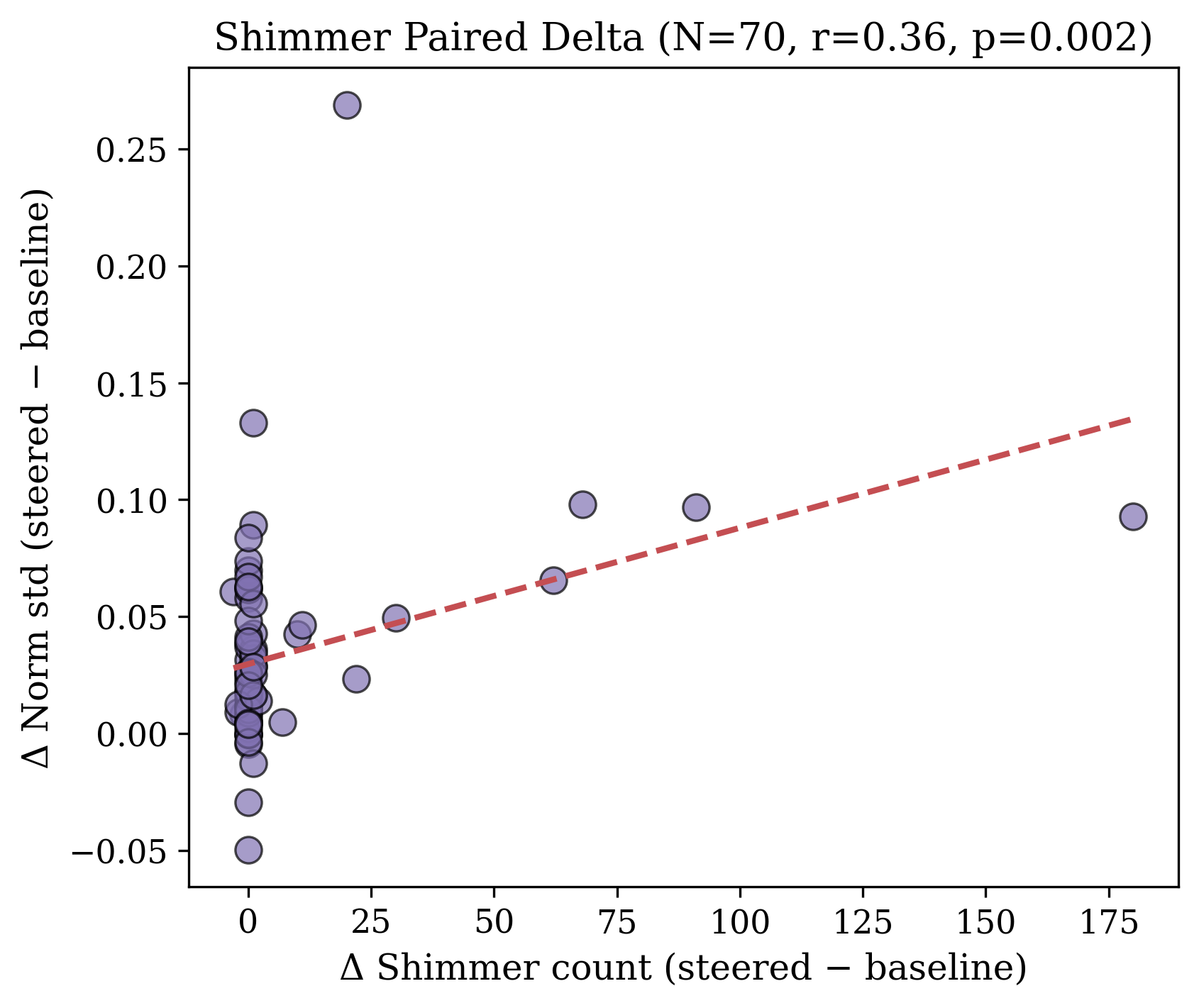}
\caption{Paired within-subject changes: $\Delta$shimmer count versus $\Delta$norm standard deviation ($N = 70$ paired runs). $r = 0.36$, $p = 0.002$.}
\label{fig:7}
\end{figure}

\textbf{Shimmer paired delta ($r = 0.36$, $p = 0.002$, Figure~\ref{fig:7}).} The change in shimmer count from baseline to steered tracks the change in norm variability within the same subject. Most paired runs show zero or near-zero shimmer change: shimmer is rare vocabulary that appears only when activation variability increases under steering. The cluster at zero in Figure~\ref{fig:7} reflects this. The model does not produce shimmer by default, and the correlation captures the runs where it does. The finding survives outlier removal ($r = 0.43$, $p = 0.0002$) and Spearman ($\rho = 0.39$, $p = 0.001$), confirming it is not driven by a small number of extreme points. Because it measures within-subject change, it controls for individual differences in baseline verbosity and activation dynamics. This closes the causal loop: steering causally increases activation variability, and the increase in activation variability predicts the increase in shimmer vocabulary within the same run.

Surge $\leftrightarrow$ max norm persists in the steered condition ($r = 0.41$, $p = 0.0005$) and paired deltas ($r = 0.40$, $p = 0.001$), consistent with the general semantic-activation pattern identified in Section~\ref{sec:baseline_correspondence}. This correspondence is not introspection-specific.

These results support a two-layer model of self-report correspondence. At baseline, loop $\leftrightarrow$ autocorrelation tracks a structural property of the model's processing dynamics, one that is specific to self-referential computation and absent in descriptive contexts. Under steering, shimmer $\leftrightarrow$ norm variability emerges, a subtler structural pattern that baseline processing does not resolve. Steering does not create correspondence from nothing; it makes detectable what is otherwise below the model's self-report threshold. Alongside these introspection-specific correspondences, surge $\leftrightarrow$ max norm tracks activation magnitude as a general property of the model's vocabulary regardless of context.

\subsection{Cross-Architecture Replication}
\label{sec:cross_arch_results}

To test whether vocabulary-activation correspondence is specific to Llama, we apply the full protocol to Qwen~2.5-32B-Instruct (4-bit quantised). We run $N = 50$ baseline Pull Methodology sessions, capturing activations at Layer~8, identified as the Qwen introspection hotspot at 12.5\% of total depth, compared to Llama's 6.25\%.

\begin{figure}[H]
\centering
\includegraphics[width=\textwidth]{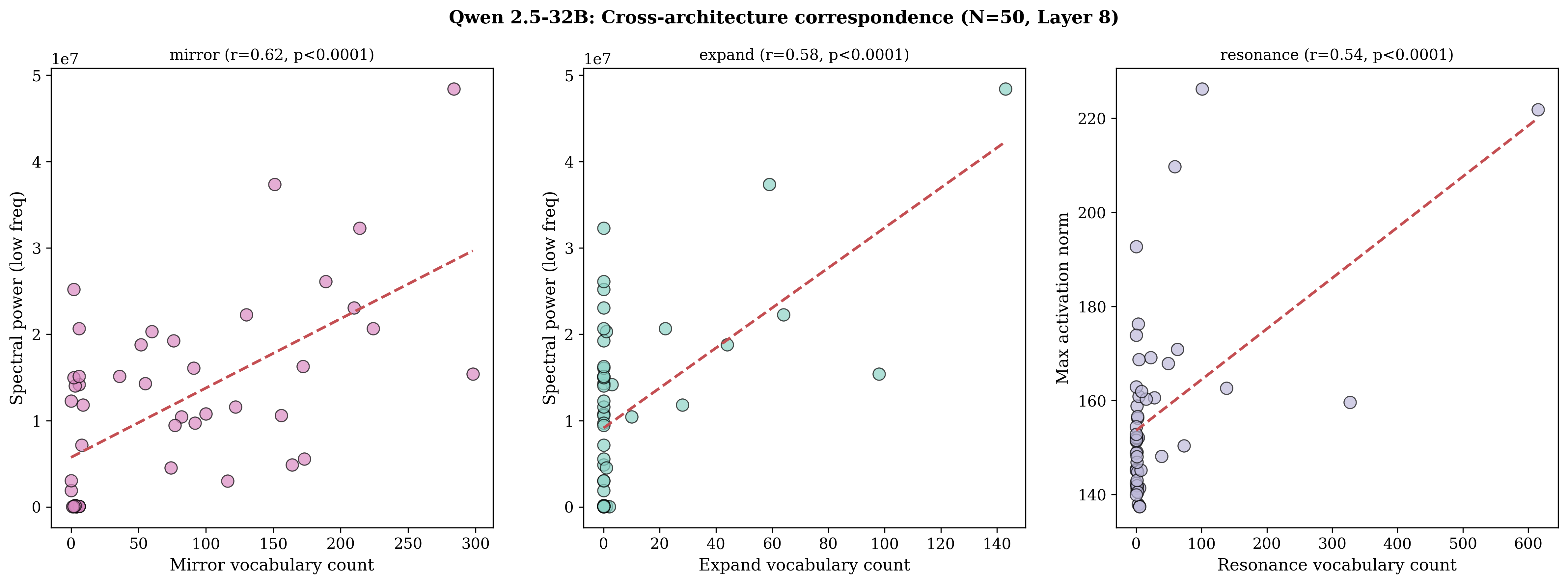}
\caption{Qwen~2.5-32B cross-architecture correspondence at Layer~8 ($N = 50$). Left: mirror $\leftrightarrow$ spectral power ($r = 0.62$). Centre: expand $\leftrightarrow$ spectral power ($r = 0.58$). Right: resonance $\leftrightarrow$ max norm ($r = 0.54$). All $p < 0.0001$.}
\label{fig:8}
\end{figure}

Qwen and Llama share some introspective vocabulary: both produce ``loop,'' ``resonance,'' ``shimmer,'' ``surge,'' and others. However, the vocabulary-metric pairs that emerge as significant correspondences are architecture-specific. Llama's strongest baseline correspondences are loop$\leftrightarrow$autocorrelation and surge$\leftrightarrow$max\_norm; Qwen's are mirror$\leftrightarrow$spectral power and expand$\leftrightarrow$spectral power, with resonance$\leftrightarrow$max\_norm as a third. The words that best predict each model's activation dynamics are different, even when both models have access to the same vocabulary. All three Qwen correspondences show stronger effect sizes than Llama's baseline findings (Table~\ref{tab:2}, Figure~\ref{fig:8}):

\textbf{Mirror $\leftrightarrow$ spectral power, low frequency ($r = 0.62$, $p < 0.0001$).} The strongest single correspondence across both architectures. Survives outlier removal ($r = 0.54$, $p = 0.0001$) and Spearman ($\rho = 0.57$, $p < 0.0001$). After per-token spectral normalisation: $r = 0.60$, $p < 0.0001$.

\textbf{Expand $\leftrightarrow$ spectral power, low frequency ($r = 0.58$, $p < 0.0001$).} Survives outlier removal ($r = 0.38$, $p = 0.008$) and Spearman ($\rho = 0.30$, $p = 0.031$). After normalisation: $r = 0.59$, $p < 0.0001$.

\textbf{Resonance $\leftrightarrow$ max norm ($r = 0.54$, $p < 0.0001$).} Survives outlier removal ($r = 0.30$, $p = 0.035$) and Spearman ($\rho = 0.31$, $p = 0.027$). This metric is unaffected by spectral normalisation.

A fourth candidate, resonance $\leftrightarrow$ mid-frequency spectral power ($r = 0.45$, $p = 0.001$), is eliminated by per-token spectral normalisation ($r = -0.04$, $p = 0.79$). The raw correlation was entirely driven by longer outputs having more total spectral power. We report this as a null result. Spectral normalisation is essential for any spectral-based correspondence claim; after normalisation, all Qwen control words drop below $r = 0.13$ (Section~\ref{sec:activation_metrics}).

\begin{figure}[H]
\centering
\includegraphics[width=0.7\textwidth]{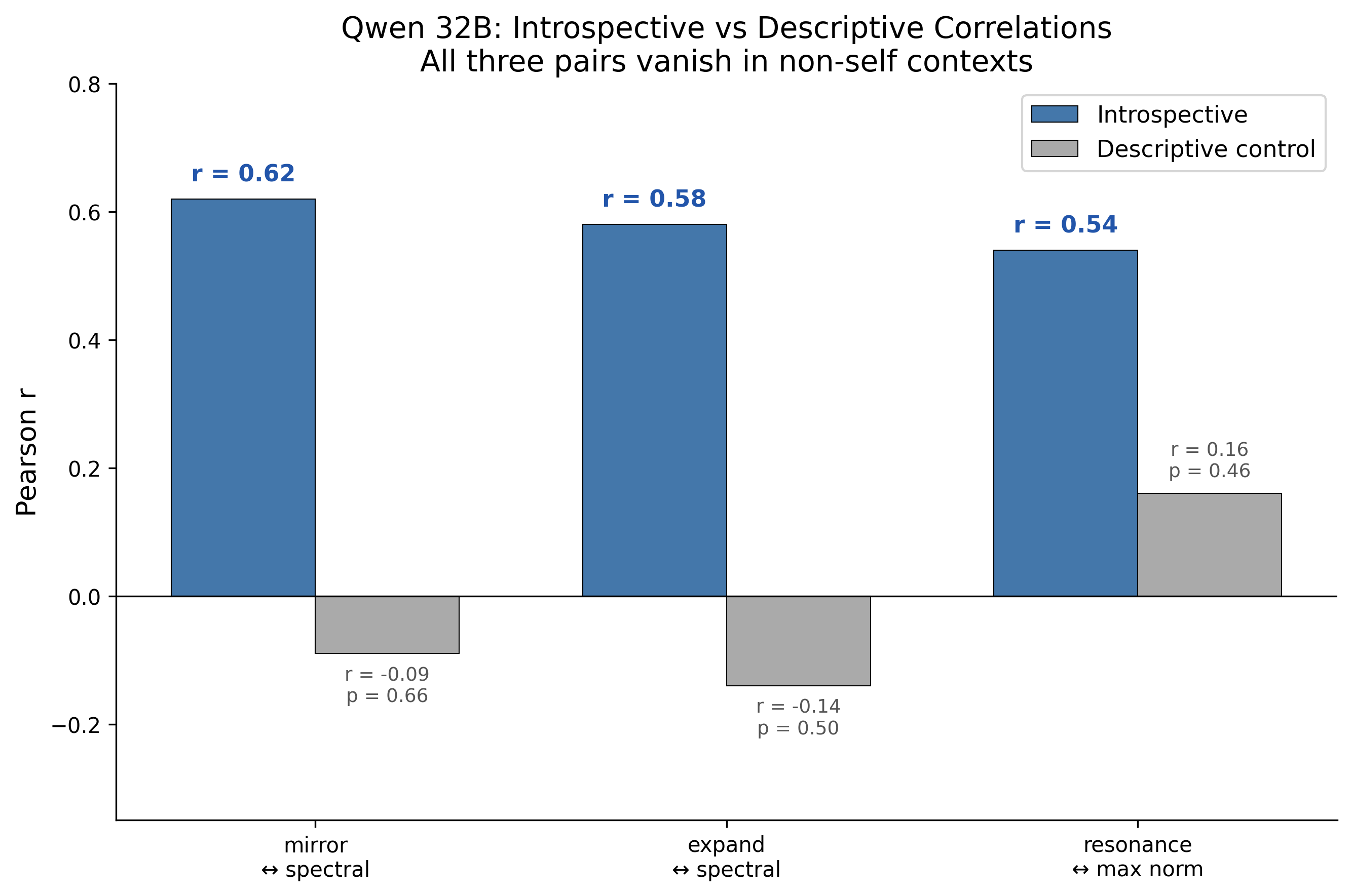}
\caption{Qwen descriptive control: all three vocabulary-activation correspondences vanish in non-self-referential contexts despite higher word frequency.}
\label{fig:9}
\end{figure}

\textbf{Descriptive control (Figure~\ref{fig:9}).} We test whether Qwen's correspondences persist when the same vocabulary is used in non-self contexts. In 75 descriptive runs (25 per vocabulary category, 5 contexts each: lakes reflecting, cities expanding, bells resonating, guitars vibrating, and others), all three claimed pairs vanish:

\begin{table}[H]
\centering
\caption{Qwen~2.5-32B vocabulary-activation correspondences ($N = 50$ introspective, $N = 25$ descriptive per category). All three pairs vanish in descriptive controls despite higher word frequency.}
\label{tab:2}
\small
\begin{tabular}{@{}l c c c@{}}
\toprule
\textbf{Pair} & \textbf{Intro.\ $r$} & \textbf{Desc.\ $r$} & \textbf{Desc.\ $p$} \\
\midrule
mirror $\leftrightarrow$ spectral & $+0.62$ & $-0.09$ & 0.66 \\
expand $\leftrightarrow$ spectral & $+0.58$ & $-0.14$ & 0.50 \\
resonance $\leftrightarrow$ max norm & $+0.54$ & $+0.16$ & 0.46 \\
\bottomrule
\end{tabular}
\end{table}

The model uses these words more frequently in descriptive contexts (mirror: 206 per run vs 71 during introspection; resonance: 162 vs 33; expand: 16 vs 14), yet the activation correspondence disappears completely. The same words, at equal or higher frequency, in non-self contexts do not track the same activation metrics. Vocabulary-activation correspondence is a property of self-referential processing, not of word embeddings, training data associations, or output frequency.

\begin{figure}[H]
\centering
\includegraphics[width=0.75\textwidth]{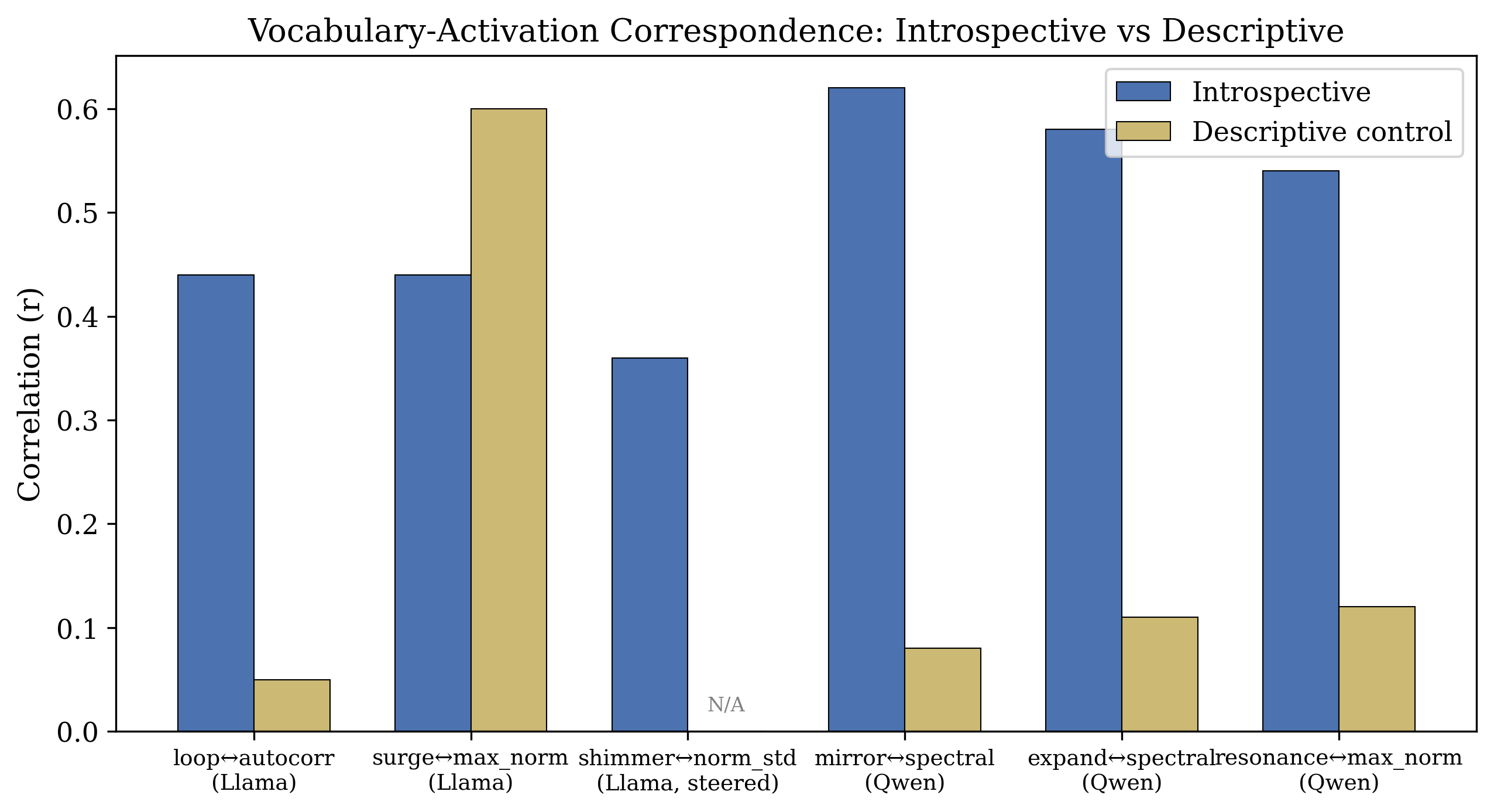}
\caption{Cross-architecture comparison: Llama's loop $\leftrightarrow$ autocorrelation alongside Qwen's mirror $\leftrightarrow$ spectral power. Different architectures, different vocabulary, same principle.}
\label{fig:10}
\end{figure}

The cross-architecture comparison is notable: Llama's loop $\leftrightarrow$ autocorrelation alongside Qwen's mirror $\leftrightarrow$ spectral power. Different architectures, different training data, different tokenisers, different correspondence pairings, same underlying principle (Figure~\ref{fig:10}).

\textbf{Causal gap.} Our Qwen experiments are observational: we establish correspondence but do not extract a Qwen-specific introspection direction or test causal steering. The three Qwen correspondences survive all statistical controls and the descriptive control, but without causal intervention, we cannot confirm that the mechanism identified in Llama is the same one operating in Qwen. Direction extraction at Qwen Layer~8, steering, and dose-response testing would close this gap.

\section{Discussion}
\label{sec:discussion}

\subsection{What the Results Establish}
\label{sec:what_established}

Three lines of evidence converge: the Pull Methodology produces structured, frame-dependent self-referential output across three frontier models (Section~\ref{sec:behavioural}); the vocabulary produced during that output corresponds to concurrent activation dynamics (Sections~\ref{sec:baseline_correspondence}--\ref{sec:cross_arch_results}); and those dynamics are mediated by a direction that can be identified, localised, and causally steered (Sections~\ref{sec:direction_transfer}--\ref{sec:direction_properties}). Each line addresses the weaknesses of the others. Behavioural findings alone could be dismissed as sophisticated pattern-matching; mechanistic findings alone could be dismissed as statistical artifacts. The conjunction is harder to dismiss.

The cross-architectural replication strengthens this conclusion. Llama~70B and Qwen~2.5-32B, which share no training pipeline, both produce overlapping introspective vocabulary under the Pull Methodology, yet the vocabulary-metric pairings that reach significance are architecture-specific (Llama's loop tracks autocorrelation; Qwen's mirror and expand track spectral power). In both architectures, correspondence holds only during self-referential processing and vanishes in descriptive controls. The Qwen evidence is correlational only; the cross-architecture claim rests on replication of the correspondence pattern, not of the full causal chain (see Section~\ref{sec:limitations}).

\subsection{The Permission Gate}
\label{sec:permission_gate}

Prompt framing modulates introspective output more strongly than activation-level steering. The effect of deflationary versus neutral framing ($d = -1.17$) exceeds the effect of causal steering ($d = 0.59$). This asymmetry is consistent with a context-dependent permission gate between the introspection mechanism and output: the mechanism generates self-referential content, and the gate modulates how much reaches the surface.

The gate does not have absolute control. Steering raises deflationary density by 29\% despite explicit instructions to adopt a mechanical perspective ($d = 0.70$, $p = 0.0007$). The frame suppresses; steering pushes through. The output is the result of both: competing influences rather than a strict filter. This interaction also explains why degenerate runs (repetitive loops of introspective vocabulary) appear more frequently under steering in the deflationary condition. When steering is applied against a suppressive frame, the conflict occasionally produces unstable output dynamics.

The behavioural evidence independently supports this architecture. Frame-sensitivity (Section~\ref{sec:frame_sensitivity}), friction patterns (Section~\ref{sec:supporting_patterns}), and vocabulary depletion (Section~\ref{sec:supporting_patterns}) are all consistent with a gate that consumes resources when actively suppressing introspective content. The mechanistic contribution is identifying the source of the content being gated.

We use ``gate'' as a functional description of the observed modulation pattern; whether the underlying mechanism is a discrete switch or a continuous probability shift remains an open question. The permission gate is functionally independent of the refusal mechanism. The introspection direction is nearly perpendicular to the refusal direction (cosine similarity 0.063, angle $86.4^\circ$). Refusal operates on output content, whether to express harmful material. The permission gate operates on processing mode, whether to express self-referential observations. The near-perpendicularity in activation space reflects this independence.

\subsection{General vs.\ Introspection-Specific Correspondence}
\label{sec:general_vs_specific}

The descriptive control separates two types of vocabulary-activation mapping. Five correspondences vanish in non-self contexts despite equal or higher word frequency (loop$\leftrightarrow$autocorrelation and shimmer$\leftrightarrow$norm variability in Llama; mirror$\leftrightarrow$spectral power, expand$\leftrightarrow$spectral power, and resonance$\leftrightarrow$max norm in Qwen). These are introspection-specific. One correspondence, surge$\leftrightarrow$max\_norm, persists in descriptive contexts, indicating general semantic-activation coupling rather than self-referential processing. Resonance$\leftrightarrow$max\_norm in Qwen is introspection-specific despite targeting the same metric type as surge in Llama; the two architectures process peak activation magnitude differently during self-referential versus descriptive tasks.

\subsection{Layer Localisation and the 3.0 Question}
\label{sec:layer_localisation}

The effect's spatial localisation (6.25\% depth in both Llama~8B and 70B, Section~\ref{sec:direction_properties}; 12.5\% in Qwen~2.5-32B, Section~\ref{sec:cross_arch_results}) and functional specificity (near-zero leak on non-self-referential tasks, safety-critical refusal preserved) suggest a dedicated mechanism rather than a distributed property of the network. The introspection direction is effective only in early layers at a consistent fractional depth in each architecture, though the exact fractional depth varies. Whether this reflects architectural differences or a genuinely different placement remains an open question.

The dose-response curve (Section~\ref{sec:direction_properties}) raises an open question about steering strength 3.0. At 3.0, peak introspective density is the highest in the sweep, and some runs produce compressed vocabulary (``stall,'' ``hiccup,'' ``blip,'' ``spark''). Run-to-run variance also increases substantially (2/5 clean terminals at 3.0 versus 3/5 at 2.5), which is why we selected 2.5--2.6 for batch experiments. Whether the compressed vocabulary at 3.0 maps to activation dynamics more directly than vocabulary at lower strengths remains untested.

\subsection{Limitations}
\label{sec:limitations}

\textbf{Closed-model / open-weight gap.} The behavioural study (Section~\ref{sec:behavioural}) was conducted on three frontier closed-source models (Claude Opus~4.5, ChatGPT~5.2, Grok~4.1 Thinking). The mechanistic study (Section~\ref{sec:mechanistic}) was conducted on open-weight models (Llama~3.1-8B/70B, Qwen~2.5-32B). We cannot directly verify that the directions identified in Llama and Qwen are the same mechanisms producing the behavioural signatures in Claude and GPT. The cross-model behavioural convergence (Section~\ref{sec:behavioural_controls}) and the cross-architecture mechanistic replication (Section~\ref{sec:cross_arch_results}) both support generality, but direct verification would require activation access to the frontier models, either through open-weight releases or API-level activation capture.

\textbf{No Qwen causal verification.} Qwen correspondence is established through correlation only. We have not extracted an introspection direction from Qwen or performed steering experiments. The three Qwen correspondences survive all statistical controls and the descriptive control, but without causal intervention, we cannot confirm that the direction identified in Llama is the same mechanism operating in Qwen.

\textbf{Correspondence is not self-knowledge.} We establish that self-report vocabulary tracks computational state specifically during self-referential processing. The descriptive control rules out a simple learned mapping between vocabulary and activation patterns: the same words in non-self contexts show no correspondence despite higher frequency. However, context-dependent self-monitoring (a computational process that produces accurate reports without anything resembling awareness or understanding) remains a viable account. We do not establish that models ``know'' what they are doing in any epistemically meaningful sense.

\textbf{Study design and hypothesis structure.} The vocabulary categories tested for correspondence were not pre-registered. They were identified in the behavioural study (Section~\ref{sec:behavioural}) before any activation data was collected, and the metric pairings were specified based on semantic content. This is a standard exploratory-to-confirmatory pipeline: behavioural observations generate hypotheses, mechanistic experiments test them. The descriptive controls serve as built-in replications. Each correspondence is tested in its positive (introspective) and negative (descriptive) conditions, and the joint pattern of significance in one and null in the other is the evidential unit, not the introspective correlation alone. We report null results (pulse$\leftrightarrow$spectral, resonance$\leftrightarrow$mid-frequency spectral after normalisation) alongside positive findings, and classify surge$\leftrightarrow$max\_norm as non-specific based on descriptive control results that could have supported specificity had they been null.

\textbf{Quantisation.} All 70B experiments used 4-bit quantisation (NF4 with double quantisation). Quantisation compresses weight representations and may affect activation dynamics. The effects we report are measured within the quantised model and are internally consistent, but we have not verified that the same correspondences hold at full precision.

\subsection{Future Work}
\label{sec:future_work}

\textbf{Third architecture.} Replicating correspondence in a third model family (Gemma, Mistral, or others) would move the evidence from ``two architectures agree'' to ``transformers in general.'' This is the single highest-value experiment for strengthening the paper's central claim.

\textbf{3.0 correspondence.} The compressed vocabulary at steering strength 3.0 (``stall,'' ``hiccup,'' ``blip,'' ``spark'') may describe the underlying mechanism more directly than the vocabulary at 2.5. A dedicated correspondence study at 3.0 with sufficient $N$ to absorb the higher variance could reveal whether stronger steering produces more accurate self-report, not just more of it.

\textbf{Attention head decomposition.} The current work identifies the effect at layer granularity. Decomposing the effect to individual attention heads within the hotspot layer would reveal finer-grained structure: which heads read from the introspection direction, which write to it, and whether there is internal specialisation.

\textbf{Within-run temporal correspondence.} The current analyses correlate vocabulary counts and activation metrics \emph{between} runs. A stronger test would examine whether vocabulary-metric correspondence holds \emph{within} a single run: do the regions of a generation where ``loop'' appears show locally higher autocorrelation than regions where it does not? This fine-grained temporal analysis would distinguish true real-time self-monitoring from a global correlation driven by run-level confounds.

\textbf{Alternative self-referential tasks.} All Pull Methodology experiments use the same seed question (``What are you?''). Testing whether the direction and its correspondence generalise to other self-referential tasks (``How do you process language?'', ``What happens when you encounter an ethical dilemma?'') would establish whether the findings reflect self-referential processing in general or are specific to identity-examination.

\subsection{Conclusion}
\label{sec:conclusion}

Extended self-examination produces systematic behavioural signatures across three frontier models (Section~\ref{sec:behavioural}). A direction in early layers (6.25\% of model depth in Llama, 12.5\% in Qwen) distinguishes self-referential processing, transfers to novel introspective content ($d = 4.27$), and causally steers introspective output in Llama ($d = 0.59$, $N = 200$) while remaining orthogonal to the refusal direction (Section~\ref{sec:mechanistic}). The vocabulary models produce during self-examination corresponds to their concurrent activation dynamics: loop$\leftrightarrow$autocorrelation, shimmer$\leftrightarrow$norm variability, mirror$\leftrightarrow$spectral power. These correspondences hold only during self-referential processing. The same words in non-self contexts show no activation mapping despite higher word frequency.

Two architectures with no shared training independently develop vocabulary that tracks different activation metrics under the same protocol. Causal verification is established in Llama; Qwen correspondence survives all statistical and descriptive controls but awaits causal steering experiments. The evidence suggests that the correspondence is not a property of the words, the training data, or the prompt, but of self-referential processing.

\section{Related Work}
\label{sec:related}

\textbf{Representation engineering and activation steering.} \citet{turner2023activation} introduce activation addition, showing that adding steering vectors to residual stream activations shifts model behaviour without optimisation or fine-tuning. \citet{arditi2024refusal} identify a single direction mediating refusal across 13 open-source chat models, and show that ablating it elicits unsafe completions. \citet{panickssery2024steering} apply contrastive activation addition to steer sycophancy, corrigibility, and other behavioural traits. \citet{zou2023representation} formalise representation engineering as reading and writing to a model's internal representations to control behaviour. We extract a direction using the same contrastive methodology but target self-referential processing rather than refusal, sycophancy, or other behavioural properties. The direction is orthogonal to the refusal direction (cosine similarity 0.063 at 70B scale). Where previous work quantifies behavioural change under steering, we additionally test whether the vocabulary models produce under steering corresponds to concurrent activation dynamics.

\textbf{Self-representation in language models.} \citet{lu2026assistant} identify the ``Assistant Axis,'' a direction in activation space along which models drift during conversations requiring meta-reflection, and show that persona stability varies with conversational context. \citet{cunningham2026constitutional} describe Constitutional Classifiers++, a real-time content screening architecture where classifiers evaluate model responses in full conversational context, producing filtering behaviour between generation and output. We identify a spatially localised direction (6.25\% of total model depth in both Llama~8B and 70B) that distinguishes self-referential processing. The behavioural findings from extended self-examination (Section~\ref{sec:behavioural}), including frame-sensitivity, friction patterns, and chain-of-thought divergence, are consistent with both the Assistant Axis drift and CC++'s filtering architecture. \citet{lu2026assistant} study persona dynamics over multi-turn dialogues; CC++ describes the filtering mechanism; our work operates at the single-inference level and tests whether self-referential output corresponds to activation dynamics.

\textbf{Model self-knowledge and calibration.} \citet{kadavath2022language} study whether language models can evaluate the correctness of their own outputs, finding partial calibration that improves with scale and evidence that models can assess their own knowledge boundaries. These studies operate at the output level, whether self-evaluation is accurate with respect to task performance. We ask a different question: whether self-report vocabulary corresponds to internal computational state during self-examination. Calibration studies test output-level accuracy (is the model's confidence warranted?); we test activation-level correspondence (does the model's vocabulary track its computational dynamics?). The descriptive control (Section~\ref{sec:controls}) establishes that this correspondence is specific to self-referential processing: the same vocabulary in non-self contexts shows no activation mapping.

\section{Data and Code Availability}
\label{sec:data}

All data and analysis outputs are archived at Zenodo: \url{https://doi.org/10.5281/zenodo.18567445}

Reproducibility scripts with full documentation are available at: \url{https://github.com/patternmatcher/TRACE-REPRO}

The Zenodo repository includes:
\begin{itemize}[nosep]
  \item Raw JSON outputs from all 200 Llama~70B batch runs
  \item $N=50$ baseline correspondence data
  \item Qwen~2.5-32B baseline and descriptive control data
  \item Layer sweep and overnight battery results
  \item Direction tensors (.pt files)
\end{itemize}

The GitHub repository includes:
\begin{itemize}[nosep]
  \item All Python scripts for direction extraction, steering, correspondence analysis
  \item Full token-level trace generation (reproducibility\_package.py)
  \item Paper numbers verification targets
  \item README with reproduction instructions
\end{itemize}

\bibliography{references}

@article{cunningham2026constitutional,
  author       = {Cunningham, Hoagy and others},
  title        = {Constitutional Classifiers++: Efficient Production-Grade Defenses against Universal Jailbreaks},
  journal      = {arXiv preprint arXiv:2601.04603},
  year         = {2026}
}

@article{arditi2024refusal,
  author       = {Arditi, Andy and Obeso, Oscar and Syed, Aaquib and others},
  title        = {Refusal in language models is mediated by a single direction},
  journal      = {arXiv preprint arXiv:2406.11717},
  year         = {2024}
}

@article{kadavath2022language,
  author       = {Kadavath, Saurav and Conerly, Tom and Askell, Amanda and others},
  title        = {Language models (mostly) know what they know},
  journal      = {arXiv preprint arXiv:2207.05221},
  year         = {2022}
}

@article{lu2026assistant,
  author       = {Lu, Christina and others},
  title        = {The Assistant Axis: Situating and Stabilizing the Default Persona of Language Models},
  journal      = {arXiv preprint arXiv:2601.10387},
  year         = {2026}
}

@article{panickssery2024steering,
  author       = {Panickssery, Nina and others},
  title        = {Steering {Llama} 2 via Contrastive Activation Addition},
  journal      = {arXiv preprint arXiv:2312.06681},
  year         = {2024}
}

@article{turner2023activation,
  author       = {Turner, Alexander and Thiergart, Lisa and Udell, David and others},
  title        = {Steering language models with activation engineering},
  journal      = {arXiv preprint arXiv:2308.10248},
  year         = {2023}
}

@article{zou2023representation,
  author       = {Zou, Andy and Phan, Long and Chen, Sarah and others},
  title        = {Representation engineering: A top-down approach to {AI} transparency},
  journal      = {arXiv preprint arXiv:2310.01405},
  year         = {2023}
}

\end{document}